\documentclass[sigconf, nonacm]{acmart}

\newcommand\vldbdoi{XX.XX/XXX.XX}
\newcommand\vldbpages{XXX-XXX}
\newcommand\vldbvolume{14}
\newcommand\vldbissue{1}
\newcommand\vldbyear{2022}
\newcommand\vldbauthors{\authors}
\newcommand\vldbtitle{\shorttitle} 
\newcommand\vldbavailabilityurl{URL_TO_YOUR_ARTIFACTS}
\newcommand\vldbpagestyle{plain} 

\usepackage{xcolor}
\usepackage{tabulary}
\usepackage{makecell}
\usepackage[capitalise]{cleveref}
\usepackage{geometry}
\usepackage[ruled,lined]{algorithm2e}

\newcommand{\algogr}{{\em GeneticRule}}
\newcommand{\algogrcf}{{\em GeneticRuleCF}}
\newcommand{\algogreedy}{{\em GreedyRuleCF}}

\newcommand{\set}[1]{\{#1\}}
\newcommand{\setof}[2]{\{#1|#2\}}
\newcommand{\defeq}{\stackrel{\mathrm{def}}{=}}
\newcommand{\inst}{\textsc{Inst}}
\newcommand{\calF}{\mathcal F}

\begin{document}
\title{Computing Rule-Based Explanations by Leveraging Counterfactuals}


\author{Zixuan Geng}
\affiliation{%
 \institution{University of Washington}
 \country{USA}
}
\email{zg44@cs.washington.edu}

\author{Maximilian Schleich}
\affiliation{%
 \institution{Relational AI}
 \country{USA}
}
\email{maximilian.schleich@relational.ai}

\author{Dan Suciu}
\affiliation{%
 \institution{University of Washington}
 \country{USA}
}
\email{suciu@cs.washington.edu}


\begin{abstract}
  Sophisticated machine models are increasingly used for high-stakes
  decisions in everyday life.  There is an urgent need to develop
  effective explanation techniques for such automated decisions.
  Rule-Based Explanations have been proposed for high-stake decisions
  like loan applications, because they increase the users' trust in
  the decision.  However, rule-based explanations are very inefficient
  to compute, and existing systems sacrifice their quality in order to
  achieve reasonable performance.  We propose a novel approach to
  compute rule-based explanations, by using a different type of
  explanation, Counterfactual Explanations, for which several
  efficient systems have already been developed.  We prove a Duality
  Theorem, showing that rule-based and counterfactual-based
  explanations are dual to each other, then use this observation to
  develop an efficient algorithm for computing rule-based
  explanations, which uses the counterfactual-based explanation as an
  oracle.  We conduct extensive experiments showing that our system
  computes rule-based explanations of higher quality, and with the
  same or better performance, than two previous systems, MinSetCover
  and Anchor.
\end{abstract}

\maketitle

\setcounter{page}{1}

\pagestyle{\vldbpagestyle}
\begingroup\small\noindent\raggedright\textbf{PVLDB Reference Format:}\\
\vldbauthors. \vldbtitle. PVLDB, \vldbvolume(\vldbissue): \vldbpages, \vldbyear.\\
\href{https://doi.org/\vldbdoi}{doi:\vldbdoi}
\endgroup
\begingroup
\renewcommand\thefootnote{}\footnote{\noindent
This work is licensed under the Creative Commons BY-NC-ND 4.0 International License. Visit \url{https://creativecommons.org/licenses/by-nc-nd/4.0/} to view a copy of this license. For any use beyond those covered by this license, obtain permission by emailing \href{mailto:info@vldb.org}{info@vldb.org}. Copyright is held by the owner/author(s). Publication rights licensed to the VLDB Endowment. \\
\raggedright Proceedings of the VLDB Endowment, Vol. \vldbvolume, No. \vldbissue\ %
ISSN 2150-8097. \\
\href{https://doi.org/\vldbdoi}{doi:\vldbdoi} \\
}\addtocounter{footnote}{-1}\endgroup

\ifdefempty{\vldbavailabilityurl}{}{
\vspace{.3cm}
\begingroup\small\noindent\raggedright\textbf{PVLDB Artifact Availability:}\\
The source code, data, and/or other artifacts have been made available at \url{https://github.com/GibbsG/GeneticCF}.
\endgroup
}

\section{Introduction}
\label{sec:intro}

We are witnessing an increased adoption of sophisticated machine
learning models in high-stakes decisions. This leads to an urgent need
for us to find some ways to make the models more explainable and
debuggable, so that we can not only ensure the fairness of machine
learning models, but also increase the public trust from human users
of these models. Due to this need, explainable machine learning has
become an important research topic.

The literature on explanation techniques is vast (e.g., 
\cite{DBLP:conf/nips/LundbergL17, DBLP:conf/kdd/Ribeiro0G16, anchor, geco, set-cover, AR, Wachter}). 
  We refer to the excellent book on interpretable machine learning for an 
overview of these techniques~\cite{molnar2022}. At a high level, 
there are two levels of explanations depending on the scope. One is
the local explanation \cite{local}, which explains the model based on
the decision on a single instance. The other one is the global
explanation \cite{global}, which explains the model as a whole; in
this paper we focus on local explanations.

One type of local explanation is the Counterfactual Explanation model,
or Actionable Recourse, which generates a counterfactual or a
``desired'' instance based on an ``undesired'' instance. Given an
instance $x$, on which the machine learning model predicts a negative,
``bad'' outcome, the counterfactual explanation says what needs to
change in order to get the positive, ``good'' outcome.  For example, a
customer applies for a loan with a bank, the bank denies the loan
application, and the customer asks for an explanation; the system
responds by indicating what features need to change in order for the
loan to be approved, for example, the response may be {\em Increase
  the Income from 500 to 700 and decrease the Open Accounts from 4 to
  3}. The semantics is that, if these features are changed
accordingly, then the machine learning model will change its
prediction from ``bad'' to ``good''.  The counterfactual provides the
user with some recommendation for what they should do in order to
change the outcome in the future.

However, in high-stakes applications like financial decisions,
counterfactual explanations may actually be misleading customers.  For
example, a different customer may have an income of 500 and 4 open
accounts yet her loan application was approved.  The reason is that
the two customers differ in other features used by the system, but
customers unfamiliar with the internals of the model will instead
perceive the decision as unfair, because they are asked to change
features that appeared to be no problem for other customers.

Counterfactual explanations appear to be insufficient for high-stakes
applications of machine learning.  For that reason, Rudin et
al.~\cite{fico_model,set-cover} argue in favor of a new kind of
explanation, called {\em rule-based explanation}. A rule is a
conjunction of predicates on the features for which the machine
learning model always generates the bad outcome.  For example a rule
could be {\em all customers who had an Income $\leq 500$ and an
  Employment-history of $\leq 10$ years were denied the load
  application}.  While this does not immediately tell the customer how
to intervene to change the outcome, it nevertheless assures her of the
fairness of the decision, because all customers with these features
had been denied.  Instead of trying to be prescriptive and instruct
the customer on what to do, rule-based explanations are descriptive in
that they provide fundamental reasons for the decision.  Rule-based
explanations are similar to the {\em anchors}, introduced by Ribeiro
et al.~\cite{anchor}, and are highly desired by financial institutions.

{\em Black-box} explanation systems compute the explanations by
  repeatedly probing the black-box classifier with inputs derived from
  instance to be explained, and from the domain of some large dataset
  of instances, which can be the data used to train the classifier, or
  some historical data of past decisions.  This leads to distinct
  computational challenges for both counterfactual and rule-based
  explanations.  A counterfactual explanation consists of some,
  ideally small, set of features, and new values for these features
  that lead to a positive outcome.  A rule-based explanation consists
  of some, ideally small, set of features, where the current values
  the instance always lead to a negative outcome, for any values of
  the other features.  In other words, a counterfactual explanation
  requires answering an existentially quantified query, while a
  rule-based explanation requires answering a universally quantified
  query.
  Finding counterfactual explanations is relatively easier, and several
counterfactual explanation systems have already been developed, such
as Mace~\cite{mace}, Geco~\cite{geco}, Dice~\cite{dice} and others,
and are capable of finding efficiently counterfactual explanations of
high quality.  In contrast, finding rule-based explanations is much
harder.  For example, the approach described by Rudin et
al. \cite{set-cover} converts the problem to a minimum set-cover
problem, whose size depends on the size of the database, then solves
it using Integer Programming.

In this paper we propose a novel approach to compute rule-based
explanations, by reducing the problem to computing counterfactual
explanations, then using an existing counterfactual system to develop
a highly efficient rule-based explanation system. We start by proving
formally that counterfactual and rule-based explanations are duals to
each other.  This means that, if a counterfactual explanation consists
of some set of features, then {\em every} rule-based explanation must
include at least one of these features. Otherwise, if the
counterfactual and rule-based explanations use disjoint sets of
features, then we are lead to a contradiction, since one explanation
asserts that by changing the values of only the first set of features
the outcome will be positive, while the other asserts that if we keep
unchanged the values of the second set of features then the outcome
will always be negative.

Using the duality theorem, we develop a new approach for computing
rule-based explanations, by using a counterfactual explanation
algorithm as a black box.  We start from a baseline consisting of a
simple algorithm for computing rule-based explanations, called
\algogr, which, given an instance $x$ with a bad outcome, searches
candidate rules using a genetic algorithm. Then, we describe two
extensions. The first extension, called Genetic Rule with
Counterfactual (\algogrcf), uses a counterfactual system to create new
candidate rules. More precisely, if a candidate rule fails to be
globally consistent, then the algorithm asks for a counterfactual
explanation to the bad outcome for $x$, but under the constraint that
none of the features already included in the rule be modified.  Each
feature changed by the counterfactual is then added to the candidate
rule, and the search continues. The second extension, called Greedy
Algorithm with Counterfactual (\algogreedy), also uses the
counterfactual explanation algorithm to extend the rule, but only
applies it to the best current candidate rule.

In order to validate a rule-based explanation one has to check whether
for all possible values of the other attributes, the outcome of the
classifier remains negative.  The property is called {\em global
  consistency}, and is very expensive to check.  A critical step in
these algorithms is the global consistency test.  To reduce its cost,
the set-cover method in~\cite{set-cover} restricts the test to
instances in the database.  In our approach, we not only check
instances in the database, but we also perform the check for all
combinations of values in the domain.  For example, suppose Alice is
denied her application, and happens to have 10 open accounts.  In
order to check the consistency of the rule ``10 open accounts always
lead to a denial'', the set-cover method checks only the customers in
the database: if all customers with 10 open accounts were denied, then
it deems the rule consistent.  However, the database may contain only
a tiny sample of customers with 10 open accounts.  In contrast, our
system checks for all combinations of all attributes, e.g. age,
income, credit-score, etc, and declares the rule consistent only if
all such combinations lead to a negative outcome.  This test is
potentially very expensive, and here is where we use the
counterfactual explanation system.  More precisely, we ask it to find
a counterfactual explanation where the features in rule are fixed, and
the others can be modified arbitrarily.  For example, we ask it to
find a counterfactual by keeping the number of open accounts equal to
10: the rule ``10 open accounts'' is globally consistent only if no
such counterfactual exists.

Finally, we conduct an extensive experimental evaluation, by
evaluating our three algorithms and comparing them to both
MinSetCover~\cite{set-cover} and Anchor~\cite{anchor}.  We found that
both MinSetCover and Anchor returned rules that are not globally
consistent. For example, MinSetCover checks consistency only on the
instances in the database and $97.4\%$ of the rules generated for
Adult dataset are not globally consistent, while Anchor almost always
returns rules with redundant predicates and also about $87.0\%$ of the
rules are not globally consistent.  A {\em redundant} predicate
  is one that can be removed from the rule and still keep it globally
  consistent; Anchor uses an multi-armed bandit approach to find
  rules, which often leads to the inclusion of redundant predicates.
In contrast, our \algogrcf\, algorithm always generates globally
consistent rules with only $12.4\%$ of the rules have redundant
predicates, and our \algogreedy\, algorithm only generates globally
consistent rules, without redundant predicates.

We note that an orthogonal approach to explanations is the
  development of {\em interpretable} machine learning models.  In
  general, simple ML models such as linear regression or rule-based
  models are considered to be interpretable.  One should not confused
  the rule-based models, as discussed
  e.g. in~\cite{DBLP:conf/kdd/LakkarajuBL16}, with rule-based
  explanations considered in our paper.  The purpose of the rule-based
  model is serve as decision mechanism, while that of a rule-based
  explanation is to provide an explanation for a decision made by some
  other, usually uninterpretable model.

\textbf{Contributions}. In summary, in this paper we make the following contributions.
\begin{enumerate}
    \item We prove the Duality Theorem between counterfactual and rule
      based explanations. Section~\ref{sec:duality:theorem}.
    \item We show how to use the Duality Theorem in order to compute
      rule-based explanations by using a counterfactual-based
      explanation system.  Section~\ref{sec:convert_counterfactual_to_rules}.
    \item We describe three algorithms: \algogr, \algogrcf, and \algogreedy\ for generating the rule-based explanations. Section \ref{sec:alg}.
    \item We conduct an extensive experimental evaluation of \algogr, \algogrcf, and \algogreedy\, algorithms, and compare them with Anchor and MinSetCover. Section \ref{sec:experiment}.
\end{enumerate}


\section{Definitions}
\label{sec:background}

Let $F_1, \dots, F_n$ be $n$ features, with domains $dom(F_1)$,
$\dots$, $dom(F_n)$, which we assume to be ordered, and let
$\inst \defeq dom(F_1) \times \dots \times dom(F_n)$.  We call an
element $x \in \inst$ an {\em instance}.  We are given a black box
classifier $C$ that, on any instance $x \in \inst$, returns a
prediction $C(x)$ within the range $[0,1]$. We assume that
$C(x) \leq 0.5$ is an ``undesired'' or ``bad'' outcome, while
$C(x) > 0.5$ is ``desired'' or ``good''. For the binary classifier, we
replace its outcomes with values $\{0,\, 1\}$, where $0$ is for
``undesired'' and $1$ is for ``desired''.  Furthermore, we assume a
database $D=\set{x_1, \ldots, x_m}$ of $m$ instances, which can be a
training set, or a test set, or historical data of past customers for
which the system has performed predictions.  For each instance in the
database we write its feature values as
$x_i = (f_{i1}, \dots , f_{in})$.
 
\subsection{Rule-based Explanation}
\label{sec:Rule-based_explanations}
Fix an instance $x_i = (f_{i1}, \ldots, f_{in}) \in D$.  A {\em rule
  component}, $RC$, relevant to $x_i$, is a predicate of the form
$F_j \leq f_{ij}$ or $F_j \geq f_{ij}$, for some feature $F_j$.  For
an instance $x \in \inst$, we write $RC(x)$ for the predicate defined
by $RC$.  In other words, if $x=(f_1, \ldots, f_n)$, then $RC(x)$
asserts $f_j \leq f_{ij}$ or $f_j \geq f_{ij}$ respectively.

A {\em rule} relevant to $x_i$ is a set of rule components,
$R = \set{RC_1, \ldots, \-RC_c}$.  We write $R(x)$ for the predicate
that is the conjunction of all rule components. The {\em cardinality}
of the rule is $c$.  Notice that, in order to assert equality for some
feature, $F_j = f_{ij}$, we need two rule components, namely both
$\leq$ and $\geq$, therefore $0 \leq c \leq 2n$.  We denote by
$\inst_{R}$ the set of all instances that satisfy $R$:
\begin{displaymath}
  \inst_{R} = \setof{x}{x \in \inst,  R(x) = 1}
\end{displaymath}

Consider an instance $x_i$ that is classified as ``undesired'',
$C(x_i) \leq 0.5$, and let $R$ be some rule.  Rudin and
Shaposhnik~\cite{set-cover} propose three simple properties that, when
satisfied, can be used to offer $R$ as explanation for the bad outcome
on the instance $x_i$:
\begin{enumerate}
\item \textbf{Relevance}: the input instance $x_i$ satisfies $R$, in
  other words $x_i \in \inst_R$.
\item \textbf{Global Consistency}: all instances $x$ in $\inst$ that
  satisfy the rule $R$ are ``undesired'': $\forall x \in \inst_R$,
  $C(x) \leq 0.5$.
\item \textbf{Interpretability}: the rule should be as simple as
  possible, in other words it should have a small cardinality.
\end{enumerate}

In this paper we consider only rules that are relevant to the instance
$x_i$, hence the first property is satisfied by definition.  Our goal
is: given $x_i$ with a bad outcome, compute one (or several) globally
consistent, interpretable rule $R$.

The {\em trivial rule} relevant to the instance $x_i$ is the rule
$R_{triv}$ that contains all $2n$ rule components relevant to $x_i$;
$R_{triv} = \set{F_1 \leq f_{i1}, F_1 \geq f_{i1}, \ldots, F_n \leq
  f_{in}, F_n \geq f_{in}}$.  In other words, $R_{triv}(x)$ asserts
that the instance $x$ has exactly the same features as the instance
$x_i$.  Since $x_i$ is undesired, $R_{triv}$ is globally consistent.
However, its cardinality is very large, $2n$, and we say that the
trivial rule is not ``interpretable''.  Instead, we seek a minimal set
of rule components that are still globally consistent.

In general, checking global consistency is computationally hard.  The
number of possible instances, $|\inst|$, is exponentially large in the
number of features, and checking all of them is intractable.  For that
purpose, the authors in~\cite{set-cover} relax the global consistency
requirement, and check consistency only relative to the database
$D=\set{x_1, \ldots, x_m}$.  We call this property \textbf{Data
  Consistency}: $\forall x_k \in D \cap \inst_R$, $C(x_k) \leq 0.5$.
Anchor~\cite{anchor} does consider global consistency, but only aims
to enforce it ``with high probability'', in other words
$\frac{|\setof{x\in \inst_R}{C(x)\leq 0.5}|}{|\inst_R|}$ should be
close to 1.  As we will see in Section~\ref{sec:experiment}, explanations
returned by both MinSetCover~\cite{set-cover} and Anchor~\cite{anchor}
often fail to satisfy global consistency.

\subsection{Counterfactual Explanation}

\label{subsec:counterfactual}

While a rule-based explanation identifies a set of features whose
values necessarily lead to an undesired outcome, a counterfactual
explanation identifies some features whose values, when updated, could
possibly lead to a desired outcome.  Formally, we fix an instance
$x_i$ with a ``bad'' outcome, and define a {\em counterfactual
  explanation} to be some other instance $x_{cf} \in \inst$ with a
``good'' outcome, $C(x_i) > 0.5$.  We often represent $x_{cf}$ by
listing only the set of features where it differs from $x_i$.

A counterfactual $x_{cf}$ is required to satisfy two properties.
First, $x_{cf}$ must be {\em feasible} and {\em plausible}
w.r.t. $x_i$.  Feasibility imposes constraints on the new values, e.g.
\texttt{income} cannot exceed (say) \$1M, while plausibility imposes
constraints on how the new values in $x_{cf}$ may differ from the old
values in $x_i$, e.g. \texttt{gender} cannot change, or \texttt{age}
can only increase, etc.  We refer to these predicates as PLAF
(plausibility/feasibility) predicates, and denote the conjunction of
all PLAF predicates by $P(x_{cf})$.  Formally, a PLAF predicate
  is a formula of the form
  $\Phi_1 \wedge \cdots \wedge \Phi_m \Rightarrow \Phi_0$, where
  $\Phi_i$ is a predicate over the features of $x_i$ and $x_{cf}$.
  One example from~\cite{geco} is
  $\texttt{gender}_{CF} = \texttt{gender}_i$, which asserts that
  \texttt{gender} cannot change; another example is
  $\texttt{education}_{CF} > \texttt{education}_i \Rightarrow
  \texttt{age}_{CF} \geq \texttt{age}_i+4$, which asserts that, if we
  ask the customer to get a higher education degree, then we should
  also increase the age by at least 4 years.  Second, we score
counterfactuals by how many changes they require over $x_i$.  Given a
distance function $dist(x,x')$ on $\inst$, the counterfactuals that
satisfy the PLAF constraints are ranked by their distance from $x_i$.

A counterfactual explanation system takes as input an instance $x_i$
with a ``bad'' outcome, a PLAF constraint $P(x)$, and a distance
function $dist(x,x')$, then returns a rank list of counterfactuals
that satisfy $P$ and are closed to $x_i$.

\subsection{Discussion}

Different types of explanations provide the users with very different
kinds of information.  For an intuition into their differences,
consider a user Bob who has applied for life insurance, and was
denied.

The SHAP score~\cite{DBLP:conf/nips/LundbergL17}, a popular form of
explanation, assigns a fraction to each feature, for example:
$\texttt{AGE}=35\%$, \texttt{BLOOD-PRESSURE} =
20\%$, \texttt{SMOKING} = 10\%$, $\ldots$.  This defines a clear
ranking of the features, but it has limited value for the end user Bob
who was denied.  We do not consider the SHAP score in this paper.

An example of a counterfactual explanation is: ``change
$\texttt{SMOKING}$ from $\texttt{true}$ to $\texttt{false}$''.  This
has a clear meaning: if Bob quits smoking, he will get approved for
life insurance.  However, if Bob's friend Charlie also smokes, yet was
approved, then Bob will feel that he was treated unfairly.

A rule based explanation looks like this: ``everybody who has
$\texttt{SMOKING} = \texttt{true}$ and
$\texttt{BLOOD-PRESSURE} \geq 140$ will be denied''.  This does not
provide Bob with any actionable advice, but it assures him of the
fairness of the decision.


\section{Duality}
\label{sec:relationship}

A rule-based explanation and a counterfactual explanation provide
quite different information to the end user.  In both cases, a good
explanation is small: a rule relevant to $x_i$ should have only a few
rule components, while a counterfactual should change the instance
$x_i$ with only a small number of features.  Several efficient
counterfactual explanation systems exist~\cite{mace,dice,geco}, but
the existing rule-based explanation systems sacrifice global
consistency for performance~\cite{set-cover,anchor}.  In this section
we prove that the two kinds of explanations are duals, and use this
property to propose a method for computing rule-based explanations by
using an oracle to counterfactual explanations.

Before we begin, we will briefly explain why the two types of
explanations have such different complexities.  Fix a small set of
features $\calF \subseteq \set{F_1, \ldots, F_n}$. These are the
  features changed by the counterfactual explanation, or defining the
  rule components of a rule-based explanation.  In either case, we
  want $\calF$ to be small, $|\calF|=k \ll n$.  For an
  illustration, in the Yelp dataset in Sec.~\ref{sec:experiment} there
  are $n=34$ features, and typical explanations involve $k=10$
  features.  Suppose we want to check whether we can construct a
counterfactual $x_{cf}$ from $x_i$ by changing only features in
$\calF$.  An exhaustive search requires $N^k$ calls to the oracle
$C(x_{cf})$, assuming all domains have size $N$.  In practice systems
like Geco~\cite{geco} sample only a few values from each domain
$dom(F_j)$; if a counterfactual is found then it returns it, otherwise
it tries a different set of features $\calF$.  Now suppose we want to
check if the rule $R$ whose rule components are given by the features
in $\calF$ is globally consistent.  Assume for simplicity that, for
each $F_j \in \calF$, we include both $F_j \leq f_{ij}$ and
$F_j \geq f_{ij}$ in $R$.  Checking global consistency requires
$N^{n-k}$ calls to the classifier, because we need to try all values
of all $n-k$ features not in $\calF$.  Sampling is no longer
sufficient.  Worse, $k$ is much smaller than $n$, which means that the
expression $N^{n-k}$ is really large.
Referring again to the Yelp dataset, the naive complexity of a
  counterfactual explanation is $N^{10}$, and of a rule-based explanation
  is $N^{24}$.
Instead, we show here how to compute rule-based
explanations by using a counterfactual explanation system as a black
box.  This is possible due to a {\em duality} that holds between the
two kinds of explanation.

\subsection{The Duality Theorem}
\label{sec:duality:theorem}
We start with a simple lemma.
  \begin{lemma} \label{lemma:duality} If $R$ is a globally consistent
    rule, and $x_{cf}$ is any counterfactual, then $R(x_{cf})$ is
    false.
  \end{lemma}
  
\begin{proof}
  By definition, if $R$ is globally consistent, then for all $x'$: if
  $R(x')$ is true then the classifier returns the ``bad'' outcome on
  $x'$, i.e. $C(x') \leq 0.5$.  Also by definition, if $x_{cf}$ is a
  counterfactual, then the classifier returns the ``good'' outcome,
  i.e. $C(x_{cf}) > 0.5$. It follows immediately that $R(x_{cf})$ must
  be false.
\end{proof}

Fix an instance $x_i=(f_{i1}, \ldots, f_{in})$ with a bad
outcome. For any other instance
  $x=(f_1, \ldots, f_n) \in \inst$, we will construct a {\em dual}
  rule $R_x$ consisting of all rule components relevant to $x_i$ that
  are {\em false} on $x$, as follows.  If $f_j > f_{ij}$ then we say
  that the rule component $F_j \leq f_{ij}$ {\em conflicts} with $x$;
  if $f_j < f_{ij}$ then the rule component $F_j \geq f_{ij}$
  conflicts with $x$. In other words, an $RC$ conflicts with $x$ iff
  it is relevant to $x_i$ and $RC(x)$ is false.  The {\em dual} of
$x$ is the rule $R_x$ consisting of all components that conflict with
$x$.  We combine the rule components in the duals with $\vee$ instead
of $\wedge$.  For a simple example, if
$x_i = (F_1=10, F_2=20, F_3=30)$ and $x=(F_1=5, F_2=90, F_3=30)$ then
$R_x = (F_1 \geq 10 \vee F_2 \leq 20)$. We prove:


\begin{theorem}[Duality] \label{th:duality} Fix a globally consistent
  rule $R$ relevant to $x_i$, let $x_{cf,1}, \ldots,x_{cf,k}$ be
  counterfactual instances, and let
  $R_{x_{cf,1}}, \ldots, R_{x_{cf,k}}$ be their duals.  Then $R$ is a
  set cover of $\set{R_{x_{cf,1}}, \ldots, \-R_{x_{cf,k}}}$.  In other
  words, for every counterfactual $x_{cf,m}$ the rule $R$ contains at
  least one rule component that conflicts with $x_{cf,m}$.
  Conversely, fix any counterfactual $x_{cf}$, and let
  $R_1, \ldots, R_k$ be globally consistent rules.  Then the dual
  $R_{x_{cf}}$ is a set cover of $\set{R_1, \ldots, R_k}$.  
\end{theorem}

\begin{proof}
  Assume the contrary, that $R$ and $R_{x_{cf,m}}$ do not have any
  common rule component.  Then $R(x_{cf,m})$ is true, which
  contradicts Lemma~\ref{lemma:duality}.  The converse is shown
  similarly: if $R_{x_{cf}}$ is disjoint from some rule, say $R_j$,
  then $x_{cf}$ satisfies the rule $R_j$, contradicting
  Lemma~\ref{lemma:duality}.
\end{proof}

The theorem says that globally consistent rules and counterfactuals
are duals to each other.  We will exploit the first direction of the
duality, and show how to use counterfactuals to compute efficiently
globally consistent rules.

\subsection{Using the  Duality Theorem}
\label{sec:convert_counterfactual_to_rules}

We now describe how to use a counterfactual explanation system to
compute a relevant, globally consistent, and informative rule $R$
for an instance $x_i$.  This is the key part of the algorithms
we proposed in Section \ref{sec:genetic_rule_with_gego} and
\ref{sec:gego_greedy}.

Theorem~\ref{th:duality} already implies a naive algorithm for this
purpose.  Use a counterfactual system to compute {\em all}
counterfactuals $x_{cf,1}, \ldots, x_{cf,m}$ for $x_i$,
construct $S \defeq \set{R_{x_{cf,1}}, \ldots, R_{x_{cf,m}}}$ the set
of all their duals, and output all minimal set covers $R$ of $S$.
Each set covering $R$ of $S$ is a globally consistent rule, because,
otherwise there exists an instance $x$ such that $R(x)=1$ and
$C(x)>0.5$.  This implies that $x$ is a counterfactual for
  $x_i$, but is not among $x_{cf,1}, \ldots, x_{cf,m}$ (because it
disagrees with each $x_{cf,j}$ on at least one feature), contradicting
the assumption that the list of counterfactuals was complete.
However, we cannot use this naive algorithm, because counterfactual
systems rarely return the complete list of counterfactuals.

Our solution is based on computing the rule $R$ incrementally.
Starting with $R=\emptyset$, we increase $R$ with one rule component
at a time, until it becomes globally consistent, as follows.  Assume
$R$ is any rule relevant to $x_i$, and suppose that $R$ is not
globally consistent.  We proceed as follows.

\textbf{Step 1} Construct the predicate $R(x')$ associated with the rule
$R$; we will use it as a PLAF predicate in the next step.

\textbf{Step 2} Using the counterfactual explanation system,
find a list of counterfactuals $x_{cf,1},\ldots, x_{cf,k}$ for
  $x_i$ that satisfy the PLAF predicate, i.e.  $R(x_{cf,j})=1$ for
all $j=1,k$.  The number $k$ is usually configurable, e.g. $k=10$.  If
no such counterfactual is found, then $R$ is globally consistent.

\textbf{Step 3} For each $j=1,k$, compute the dual $R_{x_{cf,j}}$ of
each counterfactual $x_{cf,j}$, i.e. the set of all rule components
that conflict with $x_{cf,j}$.  We notice that $R_{x_{cf,j}}$ is
disjoint from $R$, because $x_{cf,j}$ satisfies the PLAF $R(x)$.  Let
$S = \set{R_{x_{cf,1}}, \ldots, R_{x_{cf,k}}}$ be the set of all the
dual rules.

\textbf{Step 4} For each minimal set that covers $R_0$ of $S$, construct the
extended rule $R \cup R_0$, and repeat the process from Step 1.

We note that our use of PLAF rules differs from their original
  intent.  Rather than constraining the counterfactual, we use them to
  check if the rule candidate $R$ is globally consistent, and, if not,
  then to extend it.

\begin{example}
  We illustrate with a simple example.  Consider a customer $x_i$ with
  the following features:
  \begin{align*}
x_i =  & (\texttt{Age}=50,\texttt{AccNum}=4,\texttt{Income}=500,\texttt{Debt}=10k)
  \end{align*}
  Suppose the customer was denied the loan application, and we are
  computing a rule-based explanation for the denial.  Our current
  candidate rule $R$ is:
  \begin{align*}
      R = & (\texttt{Age}\leq 50)\wedge (\texttt{AccNum}\geq 4)
  \end{align*}
  However, the rule is not globally consistent.  We ask the
  counterfactual explanation system for counterfactuals that satisfy
  the PLAF defined by the rule $R$, and obtain these two results.  We
  highlight in red the features where they differ from $x_i$:
 \begin{align*}
   x_{cf,1}= & (\texttt{Age}=50,{\color{red}\texttt{AccNum}=5},{\color{red}\texttt{Income}=900},\texttt{Debt}=10k)\\
   x_{cf,2}= & (\texttt{Age}=50,\texttt{AccNum}=4,{\color{red}\texttt{Income}=600,\texttt{Debt}=2k})
 \end{align*}
 The first counterfactual says that if the customer increased her
 income to 900, then she would be approved, even if she had 5 accounts
 open.  The second counterfactual says that if she increases her
 income to 600 {\em and} decreases her debt to 2k then she would be
 approved.  The dual sets are:
\begin{align*}
  R_{x_{cf,1}} = & (\texttt{AccNum} \leq 4) \vee (\texttt{Income}\leq 500) \\
  R_{x_{cf,2}} = & (\texttt{Income} \leq 500) \vee (\texttt{Debt}\geq 10k)
\end{align*}
There are two minimal set covers, namely $\texttt{Income} \leq 500$
and $(\texttt{AccNum}\leq 4)\wedge(\texttt{Debt}\geq 10k)$.  We expand
$R$ with each of them and continue recursively.  More precisely, the
algorithm continues with:
\begin{align*}
  R_1 = & (\texttt{Age}\leq 50) \wedge (\texttt{AccNum} \geq 4) \wedge (\texttt{Income} \leq 500)\\
  R_2 = & (\texttt{Age} \leq 50) \wedge (\texttt{AccNum} = 4) \wedge (\texttt{Debt}\geq 10k)
\end{align*}
Suppose both are globally consistent.  Then we will choose $R_1$ as an explanation, because it is more informative: its cardinality is 3,
while the cardinality of $R_2$ is 4 (because $\texttt{AccNum}=4$
represents two rule components).  We tell the customer: {\em
  ``everybody 50 years old or younger, with 4 or more open accounts,
  and with income 500 or lower is denied the loan application''.}
\end{example}

\section{Algorithms}
\label{sec:alg}

We have shown  in the previous section that the Duality Theorem leads to
a method for computing rule-based explanations by using a
counterfactual-based explanation as an oracle.  In this section we
apply this method to derive a concrete algorithm.  More precisely, we
describe  three algorithms: 

\begin{description}
\item[\algogr:] This is a base-line algorithm, which explores the
  space of rule-based explanations using a genetic algorithm.  It does
  {\em not} use counterfactuals;
\item[\algogrcf:] This algorithm extends \algogr\ by using an oracle
  call to a counterfactual explanation system in order to generate and
  validate the rule-based explanations;
\item[\algogreedy] This algorithm replaces the genetic search
  with a greedy search: we greedily expand only the rule with the smallest 
  cardinality in the population, using the counterfactual explanation system as an oracle.
\end{description}

For \algogr\ and \algogrcf\ we have chosen a genetic algorithm, which is a meta-heuristics
for constraint optimization based on the process of natural selection. First, it defines an
initial population of candidates. Then, it repeatedly selects the fittest candidate in the
population and generates new candidates by changing and combining the selected candidates
(called mutation and crossover).  It stops when a certain criteria is met, e.g. it finds a
specified number of solutions.  We chose a genetic algorithm because (1) it is easily customizable to our problem of finding rule explanations, (2) it
seamlessly integrates counterfactual explanations to generate and verify rules, (3) it does
not require any restrictions on the underlying classifier and data, and thus is able to
provide black-box explanations, and (4) it returns a diverse set of explanations, which may
provide different rules that can give user more information.

Both \algogrcf\ and \algogreedy\ are based on the ideas in
Section~\ref{sec:convert_counterfactual_to_rules}: use a
counterfactual explanation oracle to build up the globally consistent
rules efficiently and to verify whether the generated rules are
consistent or not.

In the remainder of this section we describe the algorithms in detail:
\algogr\ in Section~\ref{sec:genetic_rule}, \algogrcf\ in Section
\ref{sec:genetic_rule_with_gego}, \algogreedy\ Section
\ref{sec:gego_greedy}.  Finally, in Section~\ref{sec:selectFittest} we
describe the fitness scoring function that we used for the candidate
selection.

\begin{algorithm}[t]\small
\caption{Pseudo-code of \algogr: \\\textbf{explain}(instance $x$, classifier $C$, dataset $D$)}
\label{alg:genetic_rule}
\SetKwBlock{Beginn}{beginn}{ende}
\SetKwRepeat{Do}{do}{while}%
$\text{POP} = [ \{F_1 \leq f_{i1} \},  \{F_1 \geq f_{i1} \} \ldots  \{F_n \leq f_{in} \}, \{F_n \leq f_{in} \}]$ \\
 \Do{$(\exists R \in \text{TOPK} : \ !\textbf{consistent}(R, D, s)) \lor (\text{TOPK} \cap \text{CAND} \neq \emptyset)$}{
 $ \text{CAND} = \textbf{crossover}(\text{POP}, c) \cup \textbf{mutate}(\text{POP}, m)$ \\
 $ \text{POP} = \textbf{selectFittest}(x,  \text{POP} \cup \text{CAND}, C, D, q, s)$ \\
 $\text{TOPK} = \text{POP}[1 : k]$
 }
 \textbf{return} TOPK 
\end{algorithm}

\subsection{\algogr} 
\label{sec:genetic_rule}

\algogr\, is our ``naive'' algorithm which generates rules using a genetic algorithm. The
pseudo-code is shown in Algorithm \ref{alg:genetic_rule}. The inputs are an
instance $x$, the classifier $C$, and a dataset $D$. The output is a set of rules that
explain instance $x$ for classifier $C$. In addition, there are five integer
hyperparameters: $q>0$ represents the number of rules kept after each iteration, $k \leq q$
is the number of rules that the algorithm returns to the user, $s$ is the number of samples
taken from $\inst$ to check for global consistency, $m$ and $c$ specify the number of new
candidates that are generated during mutation and respectively crossover. For instance, we
use the following settings in most of the experiments: $q=50$, $k=5$, $s=1000$, $m=3$, $c=2$. We
refer to Sec.~\ref{sec:experiment} for an explanation why we chose these settings.

\algogr\ first computes the initial population of rule candidates. We define the
initial population to be all possible rule candidates with exactly one rule component. The
initial candidates are likely not valid and consistent rules. Thus, \algogr\ repeatedly
applies \textbf{mutate} and \textbf{crossover} to generate new candidates, computes the
fitness score (via \textbf{selectFittest}) for each candidate, and then selects the $q$
fittest candidates for the next generation. This process is repeated until we find $k$
candidates that are consistent on both the dataset $D$ as well as $s$ samples from the more
general $\inst$ space. We further check that the top-$k$ candidates are not in the set 
of new generated candidates CAND, which means that they were stable for at least one
generation of the algorithm.

The \textbf{mutate} operator generates $m$ new rule
candidates for each candidate $R \in \text{POP}$. First, the operator finds the set $S$ of all rule components that are not part
of $R$. Then, it generates each new candidate by sampling (without replacement) a
single component from $S$ and adding it to $R$. Adding a single component at a time keeps 
the cardinality of the rules low and makes it less likely to introduce redundant rule
components.

The operator \textbf{crossover} generates $c$ new candidates for each pair of candidates,
$R_i$ and $R_j$. First, we compute the set $S = R_i \cup R_j$ of all rule components in
$R_i$ and $R_j$. Then, we randomly sample $t$ components from $S$ to form a new candidate.
We use $t = max(|R_i|,\, |R_j|)+1$, in order to keep the cardinality of the new candidate
low. We repeat this sampling process $c$ times to generate $c$ new candidates for every pair
of candidates in the population.

After generate new candidates via mutation and crossover, the \textbf{selectFittest}
operator calculates the fitness score of each candidate in POP, sorts the candidates by in 
descending order of their fitness scores, and returns the top $q$ candidates. We give more
details on the fitness scores in Sec. \ref{sec:selectFittest}.

We note that \algogr\ cannot guarantee that the returned rules are globally consistent.
Since we are only able to check global consistency for a sample of $\inst$,
we can only guarantee that the rules are data consistent and are likely to be globally
consistent.

\subsection{\algogrcf}
\label{sec:genetic_rule_with_gego}
\begin{algorithm}[t] 
\caption{Pseudo-code of \algogrcf: \\\textbf{explain}(instance $x$, classifier $C$, dataset $D$)}
\label{alg:genetic_rule_geco}
\SetKwBlock{Beginn}{beginn}{ende}
\SetKwRepeat{Do}{do}{while}%
$\text{POP} = [ \{F_1 \leq f_{i1} \},  \{F_1 \geq f_{i1} \} \ldots  \{F_n \leq f_{in} \}, \{F_n \leq f_{in} \}]$ \\
$\text{POP} = \text{POP} \cup \textbf{ CFRules}(\text{[\{\ \}]}, x, C, D)$\\
 \Do{$!\text{topk\_consistent} \lor (\text{TOPK} \cap \text{CAND} \neq \emptyset)$}{
  $\text{CAND} = \textbf{crossover}(\text{POP}, c) \cup \textbf{mutate}(\text{POP}, m)$\\
  $\text{CAND} = \text{CAND} \cup \textbf{CFRules}(\text{POP}, x, M, D)$\\
  $\text{POP} = \textbf{selectFittest}(x,  \text{POP} \cup  \text{CAND}, C, D, q, s)$ \\
 $\text{TOPK} = \text{POP}[1 : k]$\\
 $\text{topk\_consistent} = (\forall R \in \text{TOPK} : \textbf{consistent}(R, D, s) \;\land\; \textbf{consistentCF}(R, x, C, D))$\\
 }
 $\textbf{return} \text{ TOPK}$
\end{algorithm}

We next explain \algogrcf, which extends \algogr\ with a counterfactual explanation model to
generate and verify rule candidates. The pseudocode is provided in Algorithm
\ref{alg:genetic_rule_geco}. 

The main extension to \algogr\ is the \textbf{CFRules}
function. It takes as input a set of rule candidates and generates new candidates by
computing the counterfactual explanations for each input candidate. As outlined in
Sec.~\ref{sec:convert_counterfactual_to_rules}, this process involves multiple steps:
it computes the PLAF predicates for a given input candidate, then it computes the
counterfactual explanation for this candidate, and finally returns the dual of the
counterfactual explanation as a new candidate. We use \textbf{CFRules} to extend both the
initial population as well as the candidate pool in the main loop of the genetic algorithm. 

We further use the counterfactual model in the \textbf{consistentCF} function, which
verifies that the top rule candidates are globally consistent. For a given candidate, the
function runs the counterfactual model to generate a counterfactual example. If no such
counterfactual example can be found, we conclude that the rule is globally consistent. As a
consequence, \algogrcf\, provides higher global consistency guarantees than \algogr.

\emph{Performance optimizations.} Since calling the counterfactual explanation model is
expensive, we only run \textbf{CFRules} once for every three iteration or when all top-k
candidates are marked as data consistent. This setting gave us the best performance
improvements with minimal effects on the generated rules in our experiments. 

We further cache whether or not we were able to generate counterfactuals for each rule
candidate. This ensures that we only need to run the counterfactual model once per
candidate, and not multiple times for  \textbf{CFRules} and \textbf{consistentCF}. 

The algorithm has an optional post process stage (not shown in the pseudo code), 
to ensure that the returned rules have no redundant components. For each returned rule, 
we remove one rule component at a time, and check if the rule without this component is
still verified by \textbf{consistentCF}. If so, the removed component is redundant and 
can be removed from the rule. We repeat this process until the returned rules do not have 
any redundant component.

\subsection{\algogreedy}
\label{sec:gego_greedy}
\begin{algorithm}[t]
\caption{Pseudo-code of \algogreedy: \\explain (instance $x$, classifier $C$, dataset $D$)}
\label{alg:geco_greedy}
\SetKwBlock{Beginn}{beginn}{ende}
\SetKwRepeat{Do}{do}{while}%
$\text{POP} = \textbf{CFRules}(\text{[\{\ \}]}, x, C, D)$\\
$\text{POP} = \textbf{sortByCardinality}(\text{POP})$\\
 \While{$ (!\textbf{consistentCF}(\text{POP[1]}, x, C, D)) $}{
 \tcp{get and remove the top candidate from POP} 
 $\text{top\_cand} = \textbf{pop}( \text{POP} ) $ \\ 
 \tcp{generate new candidates only for top\_cand}
 $ \text{CAND} = \textbf{CFRules}([\text{top\_cand}], x, C, D)$ \\
 $\text{POP} = \textbf{sortByCardinality}(\text{POP} \cup  \text{CAND})$\\ 
 $\text{POP} = \text{POP[1 : q]}$\\}
 $\textbf{return} \text{ POP[1]}$
\end{algorithm}

\algogreedy\ does not use a genetic algorithm, but instead repeatedly utilizes the underlying counterfactual explanation model to greedily find rule candidates with small cardinality. The pseudocode is provided in Algorithm \ref{alg:geco_greedy}.  

\algogreedy\ generates the initial population by running \textbf{CFRules} on the empty rule candidate, and maintains the population sorted in increasing order with respect to the cardinality of the rule candidates. Then, the algorithm repeatedly takes the candidate with the smallest cardinality, generated new candidates by \textbf{CFRules} on this candidate, removes the considered candidate from and adds the new candidates to the population. The algorithm stops when the candidate with the smallest cardinality is found to be consistent by \textbf{consistentCF}.

In each iteration, \algogreedy\ removes the inconsistent rule candidate with the smallest cardinality, and adds candidates which have cardinalities strictly larger than the removed one. Therefore, the cardinality of the rules in the population are monotonically increasing. The algorithm is guaranteed to terminate with an consistent rule, since candidates can have at most $2n$ rule components, where $n$ is the number of variables in $D$.

\subsection{Fitness Score Function}
\label{sec:selectFittest}

We describe the fitness score that is used to rank  the rule
candidates in the  \textbf{selectFittest} function. For a given rule,
the fitness score is based on its degree of consistency and its
cardinality (a proxy of interpretability).
We define three degrees of consistency:
\begin{enumerate}
    \item The rule failed  data consistency ($FDC$): it violates instances in the database $D$; 
    \item The rule failed global consistency($FGC$): it is data
      consistent (satisfies all instances in the dataset $D$), but
      fails for some instances in $\inst$;
    \item The rule is globally consistent ($GC$): The rule is
      consistent for both the dataset $D$ as well as the instances
      from $\inst$.
\end{enumerate}

The fitness score of a rule $R$ is defined as follows.  Suppose the
database has $m \defeq |D|$ instances, each with $n$ features.  Let
$VD$ denote the number of instances in $D$ that violate $R$.  If
$VD=0$, then we sample $s$ instances from $\inst$ and denote by $VS$
the number of samples that violate $R$.  The fitness score $score(R)$
is:
\[
    score(R)= 
\begin{cases}
    0.25 \times (1 - \frac{|R|}{2 \cdot n}) + 0.25 \times (1 - \frac{VD}{m}),& FDC\\
    0.25 \times (1 - \frac{|R|}{2 \cdot n}) + 0.25 \times (1 - \frac{VS}{s}) + 0.25,   & FGC \\
    0.25 \times (1 - \frac{|R|}{2 \cdot n}) + 0.75. & GC
\end{cases}
\]

The expressions (including the coefficients 0.25, 0.75) were
  chosen to ensure that the score function always ranks candidates in
  a given level higher than the candidates in the levels below. For
  instance, the score of a global consistent rule candidate, GC, is
  always higher than those of levels FDC and FGC. If two candidates
  are in the same level, the one with smaller cardinality is ranked
  higher. This ranking ensures that we prioritize candidates that have
  better consistency guarantees.

\balance

\section{Experiments}
\label{sec:experiment}

We evaluate the three algorithms \algogr, \algogrcf, and \algogreedy\ from  Section~\ref{sec:alg}, and address the following questions: 
\begin{enumerate}
\item Do our algorithms find the correct rules, when these rules are
  known (the ground truth is known)?
    \item Do our algorithms find rules for real datasets and machine learning models, and are they globally consistent? 
    \item How does the quality of the generated rules as well as the runtime of our algorithms compare to those generated by the  state of art systems Anchor \cite{anchor} and MinSetCover \cite{set-cover}?  
    \item How effective is the integration of counterfactual explanations in the generation of the rules? 
\end{enumerate}

\begin{figure*}[t]
\minipage{0.24\textwidth}
    \centering
    \textbf{\;\; Cardinality of Classifiers = 2}
  \includegraphics[width=\linewidth]{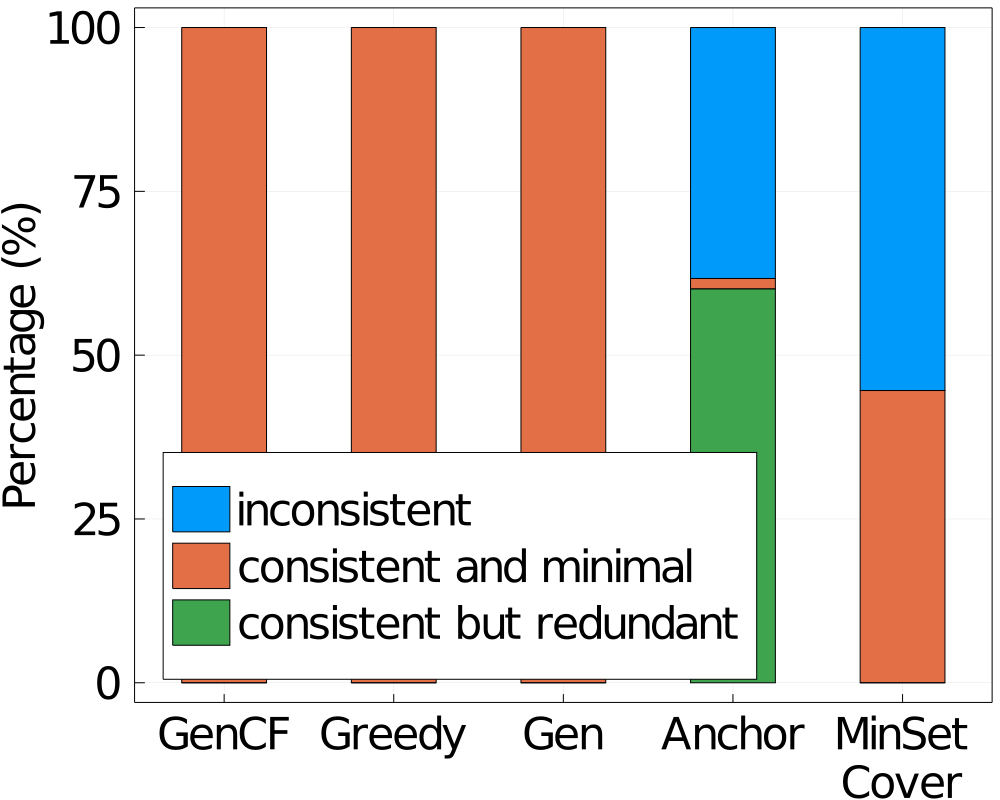}
\endminipage\hfill
\minipage{0.24\textwidth}
  \centering
  \textbf{\;\; Cardinality of Classifiers = 4}
  \includegraphics[width=\linewidth]{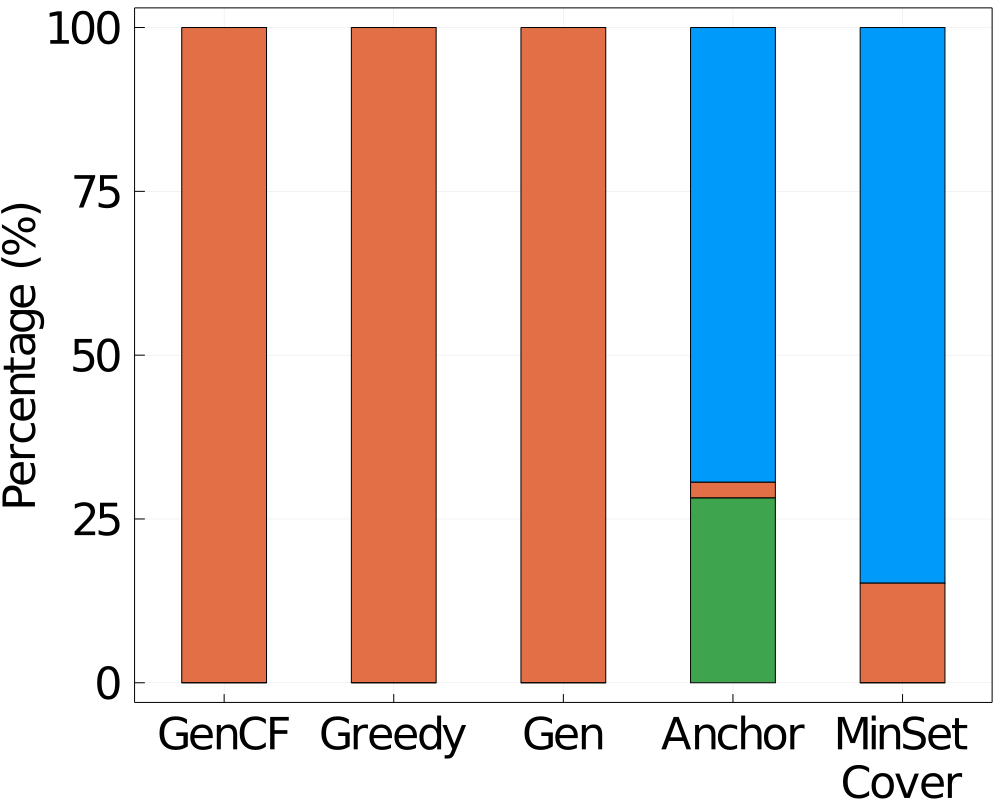}
\endminipage\hfill
\minipage{0.24\textwidth}%
  \centering
  \textbf{\;\; Cardinality of Classifiers = 6}
  \includegraphics[width=\linewidth]{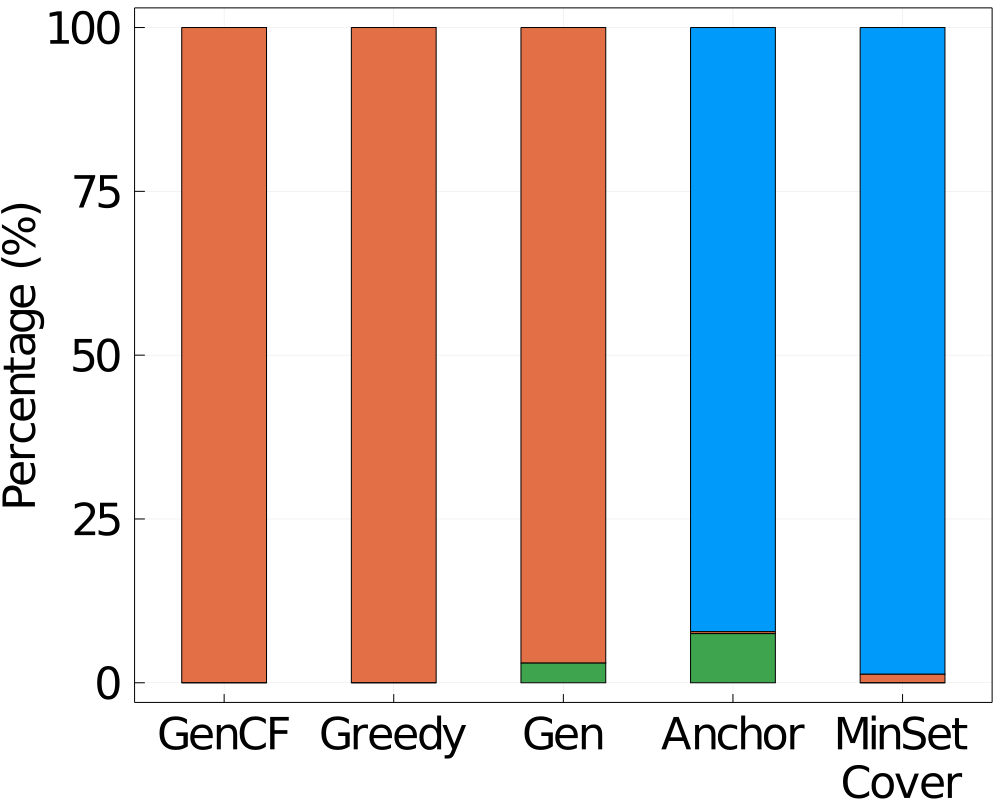}
\endminipage\hfill
\minipage{0.24\textwidth}%
  \centering
  \textbf{\;\; Cardinality of Classifiers = 8}
  \includegraphics[width=\linewidth]{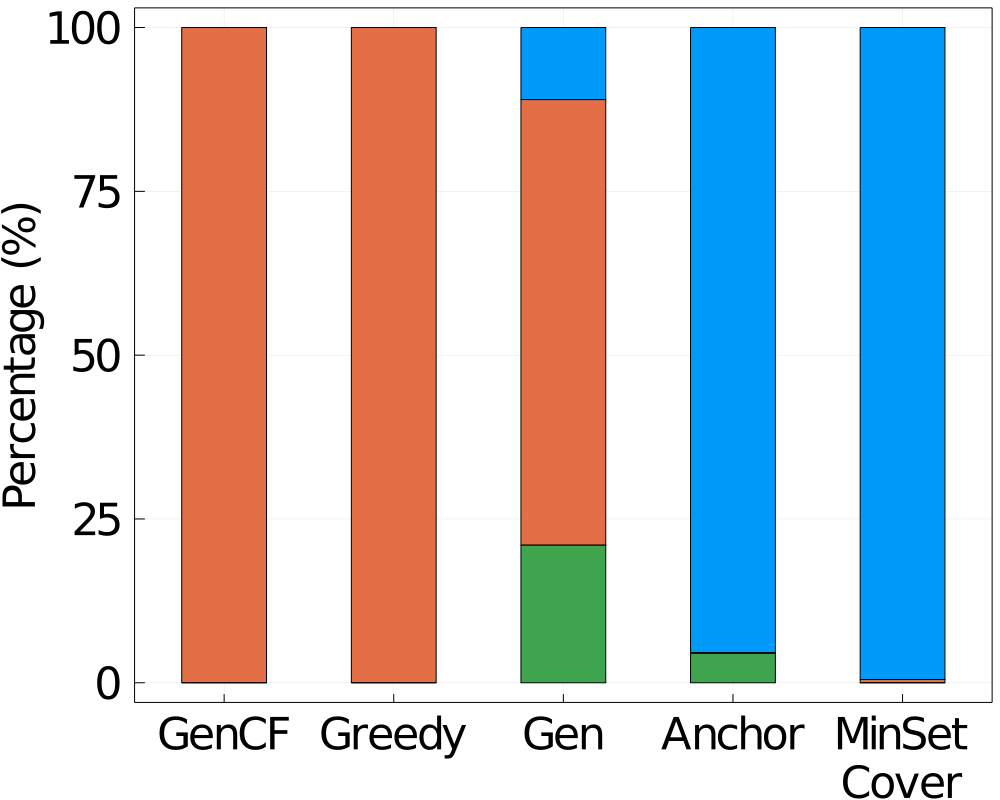}
\endminipage
\caption{The break down of the percentage of rules that are consistent and minimal, consistent with redundant components, and inconsistent for \algogr\,(Gen), \algogrcf\,(GenCF), \algogreedy\,(Greedy), Anchor, and MinSetCover over 1000 synthetic classifiers with 2, 4, 6, and 8 rule components for the Credit Dataset.}
\label{fig:breakdown_verified_syn}
\end{figure*}

\subsection{Experiment Setup}
\label{sec:exp:setup}

In this section, we introduce the datasets, systems, and the setup for all our experiments.

\textbf{Datasets and Classifiers.} We consider four real datasets: 
\begin{enumerate}
    \item Credit Dataset \cite{credit_dataset}: used to predict the default of the customers on credit card payments in Taiwan;
    \item Adult Dataset \cite{adult}: used to predict whether the income of adults exceeds $\$50$K/year using US census data from 1994;
    \item FICO Dataset \cite{fico}: used to predict the credit risk assessments.
    \item Yelp Dataset \cite{yelp}: used to predict review ratings that users give to businesses.
\end{enumerate}

\begin{table}[t]
\centering
 \begin{tabular}{|c|c|c|c|c|} 
 \hline
     & Credit & Adult & Fico & Yelp \\ 
 \hline
 \hline
 \thead{Number of Instances}
  & 30K & 45K & 10.5K & 22.4M\\ 
 \hline
  \thead{Features} & 14 & 12 & 23 & 34\\
 \hline
 \thead{Classifier Type}&\thead{Decision \\Tree}&\thead{Decision \\Tree}&\thead{Neural \\Network}&\thead{Neural \\Network} \\
 \hline
 \end{tabular}
 \vspace{1.2mm}
 \caption{Key Characteristics for each Real Dataset.}
 \vspace{-0.5cm}
 \label{tab:1}
\end{table}

\cref{tab:1} presents key statistics for each dataset and the corresponding classifiers we used. Credit and Adult are from the UCI repository \cite{uci} and are commonly used in the machine learning explanation fields. We utilize the Decision Tree classifiers for them. FICO is from the public FICO challenge, which is an Explainable Machine Learning Challenge that inspires a lot of research in this field. For the FICO dataset, we use the two-layer neural network classifier, where each layer is defined by logistic regression models. Yelp is the largest dataset we consider, and we use complex MLPClassifier with 10 layers as classifier.

In order to demonstrate whether the systems can recover the rules when
these are known (ground truth is known), we create synthetic
classifiers for the Credit dataset. The classifier is defined by a
rule, with a number of rule components, and in the experiments we
check whether the explanation algorithm can recover the rule that
defines the classifier. As usual, our algorithms do not know the
classifier, but access it as a black box.  We expect real rule-based
explanations to consists of a relatively small number of rule
components (under 10; otherwise they are not interpretable by a
typical user), so in this synthetic experiment we created classifiers
with 2, 4, 6, and 8 rule components respectively.  We repeated our
experiments for 1000 randomly generated synthetic classifiers.


In order to ensure that our evaluation is fair, we apply the same preprocessing for all the systems.

\textbf{Underlying Counterfactual Explanation Model.} To identify the best counterfactual explanation model for our algorithm, we benchmarked thirteen different counterfactual explanation models (GeCo~\cite{geco}, Actionable Recourse~\cite{AR}, CCHVAE~\cite{CCHVAE}, CEM~\cite{CEM}, CLUE\cite{CLUE},  CRUDS~\cite{cruds}, Dice~\cite{dice}, FACE~\cite{face}, Growing Spheres~\cite{gs}, FeatureTweak~\cite{ft}, FOCUS~\cite{focus}, REVISE\cite{revise}, Wachter~\cite{Wachter}) using the Carla benchmark~\cite{carla}. We report the results of this evaluation in our public GitHub repository: \url{https://github.com/GibbsG/GeneticCF}.

We find that GeCo is only one among all of these thirteen counterfactual explanation models that can robustly generate counterfactual explanations with flexible PLAF constraints and without redundant feature changes. Therefore, we decided to choose GeCo as the counterfactual explanation model for \algogrcf\,  and \algogreedy\, and to help verify the globally consistency for the rules returned by all of the considered algorithms. 

\textbf{Considered Algorithms.}
We benchmark our three algorithms, \algogr, \algogrcf, and \algogreedy, against two existing systems: Anchor \cite{anchor} and MinSetCover \cite{set-cover}. 

Anchor \cite{anchor} generates rule-based explanations (i.e. anchors) by the beam-searched version of pure-exploration multi-armed bandit problem. It starts with an empty rule. In each iteration, for each rule, Anchor (1) randomly selects $m$ possible rule components and add each of the possible components to the rule to create $m$ new rules, (2) evaluates all the rules, (3) and then selects top $n$ rule to keep for the next iteration. It stops when reaching a convergence for the rules. When evaluating whether a rule is consistent, Anchor samples $k$ instances in the whole space that are specified by the rule and considers the rule as consistent if all of those $k$ instances are ``undesired''.

MinSetCover \cite{set-cover} generates rule-based explanations using the minimum set cover problem. They consider the $m$ instances in the database as elements and the binary predicates ($\leq F_i$ or $\geq F_i$) as sets. Then, finding a minimum rule with data consistency is reduced to finding the minimum number of binary predicates that covers all those ``undesired'' instances. Therefore, MinSetCover reduces the rule-based explanations problem to finding a minimum set cover problem and solves it by Linear Programming. Note that this approach can only be applied on the historical database $D$ and does not consider instances outside this database. 

 \textbf{Parameter Choice.} As  discussed in Section \ref{sec:genetic_rule}, there are five hyperparameters in our systems: the number $k$ of rules that the algorithm returns to the user, the number $q$ of rules kept in each iteration, the number of new candidates that are generated during mutation ($m$) and crossover ($c$), the number $s$ of samples from \inst during evaluation. While $k$ depends on the user's requirements, the other parameters determine the tradeoff between the quality of the returned rules and the time it takes to return them. For instance, if we increase the number $q$ of rules kept in each iteration, it is possible that we find rule with higher quality, but we also need to evaluate and verify a larger number of candidates in each iteration of the algorithm.  We performed several pilot experiments to find the combination of hyperparameters in order to ensure that the rules are returned in a reseanonble time with acceptable quality. We found $q = 50$, $s = 1000$, $m = 3$, $c = 2$ to be best settings for the experiments with the Adult, Credit and FICO datasets. For the Yelp dataset, we use $q = 20$, $s = 5000$, $m = 3$, $c = 2$.  

For \algogrcf, we enable the optional post reduction stage, but limit it to reduce only the top rule to limit the overhead.

\textbf{Experimental pipeline.} The data is pre-processed as
  required by the classifiers: all categorical variables in the Credit
  and Yelp dataset are integer encoded, while those in the Adult are
  one-hot encoded.  We use decision tree classifer for the Credit and
  Adult dataset, and multi-layer neuron network for Fico and Yelp
  datasets. This way we explore different types of variable encodings
  and classifiers. The post-processed datasets retain the same number
  of instances (tuples) as the original data, as shown in Table
  \ref{tab:1}.  Recall that one explanation is for {\em one single
    user}, yet in order to provide explanation to one instance, the
  system needs to examine at least the entire dataset $D$, or, better,
  the entire space of instance $\inst$.  If a system returns more than
  one explanation for the user, then we retained the top explanation.
  In short, for one user (instance) we run each system to find
  rule-based explanations, and retain the top-ranked rule.  We measure
  the run time needed to find the explanation, then evaluate its
  quality.  We then repeat this process for 10,000 users (i.e. we
  compute 10,000 explanations), to get a better sense of the variance
  of our findings, for all systems.  Thus, in our experiments each
  system returns 10,000 rules, i.e. one explanation for each user.

\textbf{Evaluation Metrics.} 
We utilize the following two metrics, which are adapted from our principles of rules, to evaluate the quality of the generated rules: (1) Global Consistency: we can not find any instance that is specified by the rule and is classified as ``desired'' by the classifier; (2) Interpretability: the cardinality of the rule (i.e. the number of rule components). To be specific, we check whether there are redundant components from the return rules and whether the rule returned is minimal. While \algogr\, and \algogrcf\, generate multiple rule-based explanations for each instance, we only consider the top one rule-based explanation in our evaluation.

\textbf{Setup.} 
We implement \algogr, \algogrcf\, and \algogreedy\, algorithms in Julia 1.5.2. All experiments are run on an Intel Core i7 CPU Quad-Core/2.90GHz/64bit with 16GB RAM running on macOS Big Sur 11.6.


\begin{figure*}[htbp!]
\minipage{0.24\textwidth}
    \centering\textbf{Credit Dataset}\par\medskip
  \includegraphics[width=\linewidth]{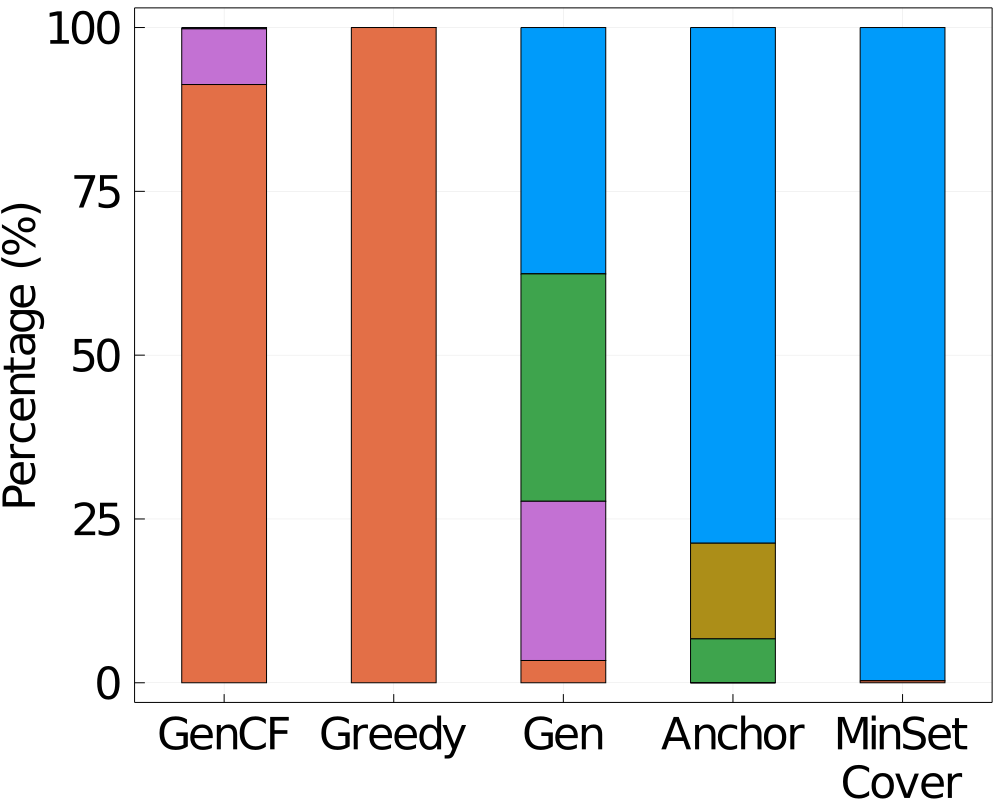}
\endminipage\hfill
\minipage{0.24\textwidth}
\centering\textbf{Adult Dataset}\par\medskip
  \includegraphics[width=\linewidth]{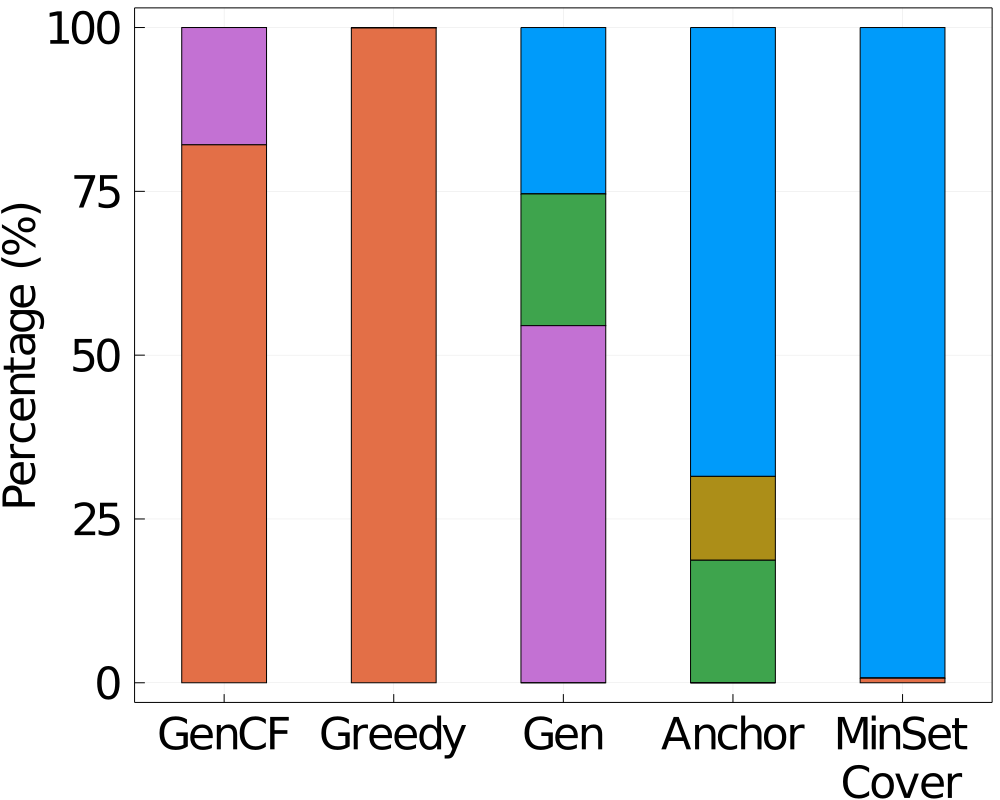}
\endminipage\hfill
\minipage{0.24\textwidth}%
\centering\textbf{Fico Dataset}\par\medskip
  \includegraphics[width=\linewidth]{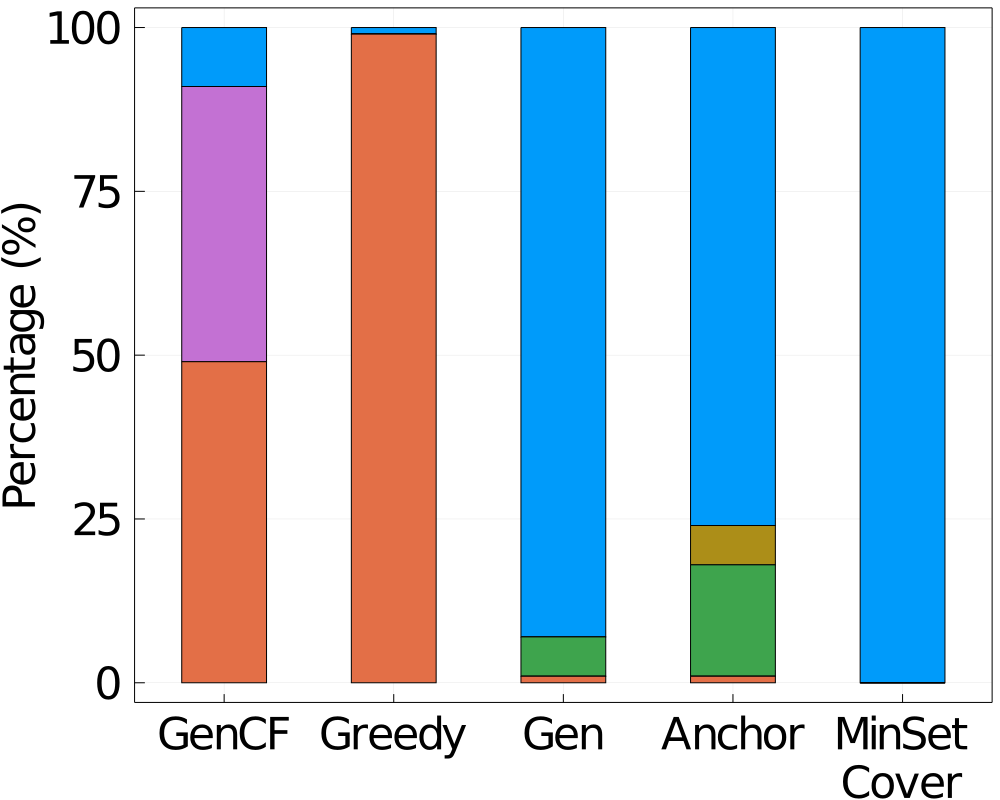}
\endminipage\hfill
\minipage{0.24\textwidth}%
\centering\textbf{Yelp Dataset}\par\medskip
  \includegraphics[width=\linewidth]{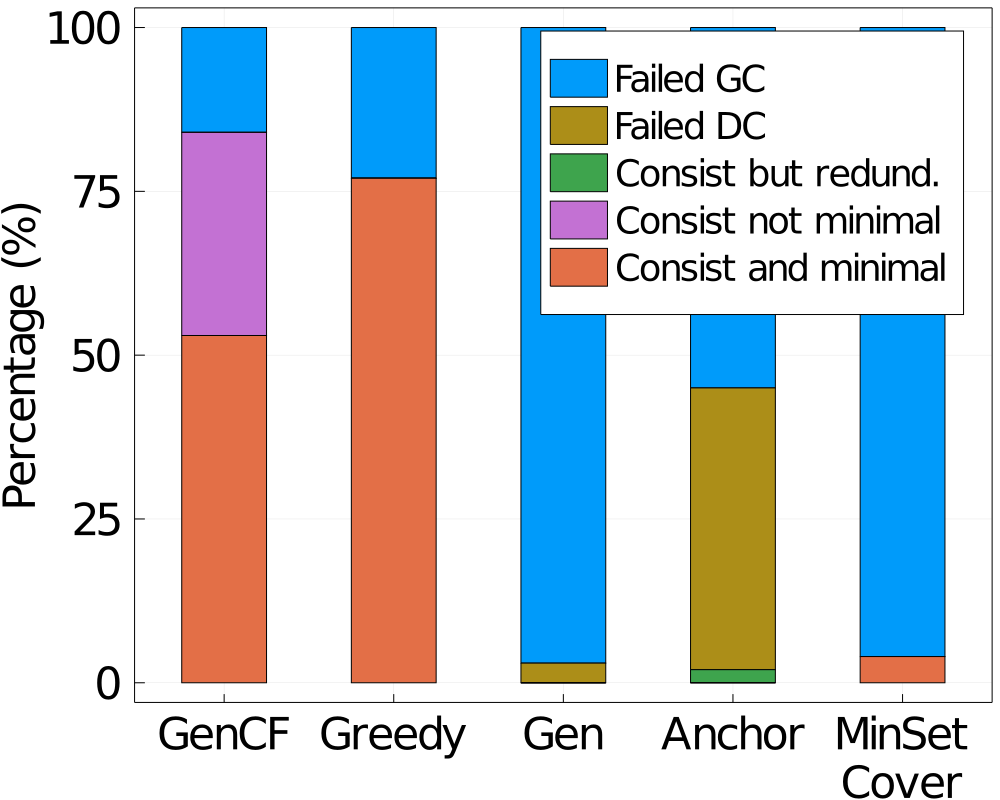}
\endminipage
\caption{The break down of the percentage of rules that are not global consistency (Failed GC), not data consistency (Failed DC), consistent with redundant components,  consistent with redundant components but not minimal (consist not minimal), and consistent and minimal for \algogr\ (Gen), \algogrcf\ (GenCF), \algogreedy\ (Greedy), Anchor, and MinSetCover. We explain 10000 instances for the Credit, Adult, and Fico dataset, and 100 instances for the Yelp dataset.} \label{fig:breakdown_verified_real}
\end{figure*}

\subsection{Quality in terms of Consistency and Interpretability}
We compare all considered algorithms in terms of the quality of generated rules on the datasets. First, we consider synthetic classifiers and then we evaluate the considered systems on real classifiers.

\textbf{Synthetic Classifiers.}  Recall that here the classifier is a
rule itself, and the task of the system is to find an explanation that
is precisely that rule.  We categorize the rule-based explanation into
three categories: (1) the rule exactly matches the classifier
(consistent and minimal); (2) the rule is consistent but has redundant
components, i.e. it is a strict superset of the classifier; or (3) the
rule is inconsistent with the classifier, i.e. it misses at least one
rule component of the classifier.  We report the percentage of rules
that fall into the three categories for each considered algorithm.

Figure \ref{fig:breakdown_verified_syn} presents the results of our evaluation on the five algorithms over $1000$ synthetic classifiers with of $2$, $4$, $6$, and $8$ rule components respectively for the Credit dataset.

\algogrcf\, and \algogreedy\, can always find the consistent and
minimal rules ($100\%$) regardless of the cardinality of the
rule. \algogr\, can always find the consistent and minimal rules when
the cardinality of the classifiers is small, but it generates some
inconsistent rules ($11\%$) for classifiers with 8 rule
components. This exemplifies the benefits of including a
counterfactual explanation system in the rule generation algorithm.

Both Anchor and MinSetCover do not always find the consistent rules even with redundancy when the cardinality of the classifers is 2 ($61.7\%$ and $44.6\%$). For larger cardinalities, they rarely generate consistent rules: only $7.8\%$ of the rules in Anchor and $1.3\%$ of the rules in MinSetCover are consistent when the cardinality of the rules bebind the classifiers are $6$. Anchor is more likely to find consistent rules than MinSetCover, but the rules generated from Anchor are mostly redundant. MinSetCover limits the cardinality of the generated rules and does not return any rules with redundant components. As a result, however, it often returns rules with fewer components than expected. In conclusion, our algorithm outperforms both Anchor and MinSetCover in terms of consistency and interpretability for the considered synthetic classifiers.

\textbf{Real Classifiers.}
For the real classifiers, we categorize each rule $R$ in one of the following five categories: 
\begin{enumerate}
\item Failed data consistency ($FDC$): there is an instance in the
  dataset $D$ where the rule fails.
\item Failed global consistency ($FGC$): all instances in $D$ satisfy
  the rule, but it fails on some an instance in $\inst$.
\item Globally Consistent ($GC$) but redundant: the rule holds on all
  instances in $\inst$, but has some redundant rule components;
\item Globally Consistent ($GC$), non-redundant, but not minimal: the
  rule is globally consistent and non-redundant, but is not of minimum
  size: there exists a strictly smaller globally consistent rule.
\item Globally Consistent ($GC$) and minimal: has the smallest number
  of rule components.
\end{enumerate}

In contrast to synthetic classifiers, we do not know the correct rules for real classifiers. We can, nevertheless, check whether a rule has redundant components by removing one of its rule components and checking if it is still consistent. If so, the rule component is redundant as the removed feature is not required. When checking the cardinalty of the miniml rules, we check all possible rule sorted by the cardinalty until finding a consistent one. Then, all consistent rules with that cardinality are considered as minimal.

Recall that our test for global consistency consists of running the
counterfactual explanations model (in our case GeCo) as a proxy: we
run GeCo subject to the constraints provided by the rule and, if GeCo
can not find a counterfactual explanation, then we conclude that the
rule is globally consistent.

Figure \ref{fig:breakdown_verified_real} shows our evaluation results for the Credit dataset with a decision tree classifier, the Adult dataset with a decision tree classifier, the Fico dataset with a multi-layer neural network classifier, and the Yelp Dataset with a MLP classifier, respectively. 

 \algogrcf\, and \algogreedy\, can always find globally consistent rules without redundancy for the Credit and Adult datasets, while they sometimes return inconsistent rules for Fico (9\% and 1\%) and Yelp (16\% and 23\%). This is because GeCo is unstable for these algorithms, and does not always find counterfactual examples. If the underlying counterfactual system is stable, \algogrcf\; and \algogreedy\; always find globally consistent rules. Further, \algogreedy\; is always able to find consistent rules that are minimal. In contrast, \algogrcf\, finds rules without redundancy, but the rules are not always minimal (the returned rule often have one or two additional rule components than the minimal rule). This shows that \algogreedy\, is the only algorithm that is able to find consistent, non-redundant, and minimal rules. 
 
\algogr\, is not guaranteed to find the globally consistent rules. For instance, for the Credit dataset 37.6\% of the generated rules by \algogr\, are not globally consistent. Unlike the synthetic classifiers, the real classifiers are more complex. And \algogr\, only uses a sample from \inst to verify for global consistency. Since the real classifiers are complex, it is possible that it finds rules that are consistent on the sample but not on the entire instance space. This further demonstrates the necessity of the counterfactual system to verify rules, as it helps to explore the instance space more broadly. 

In \algogr\, algorithm, we use a heuristic random selection to choose the rule components, which makes rules less likely to include redundant components compared with Anchor (Rules from Anchor usually have 3 or 4 redundant components, while those from \algogr\ have 1 or 2). When using GeCo in the rule-building stage, we assure that the new added rule component is necessary and not redundant. This explains the decreasing pattern in the average cardinality of the rules from Anchor and \algogr, to \algogrcf, and \algogreedy. For MinSetCover, it always finds rule-based explanations that are only data consistent, but rarely globally consistent. Thus, most of the rules returned by MinSetCover have small cardinality but are not globally consistent. This further demonstrates that our algorithms can find both consistent and interpretable rules and beat Anchor and MinSetCover on the real dataset and classifiers.


\begin{figure*}[htbp!]
\minipage{0.24\textwidth}
    \centering\textbf{Cardinality of Classifier = 2}\par\medskip
  \includegraphics[width=\linewidth]{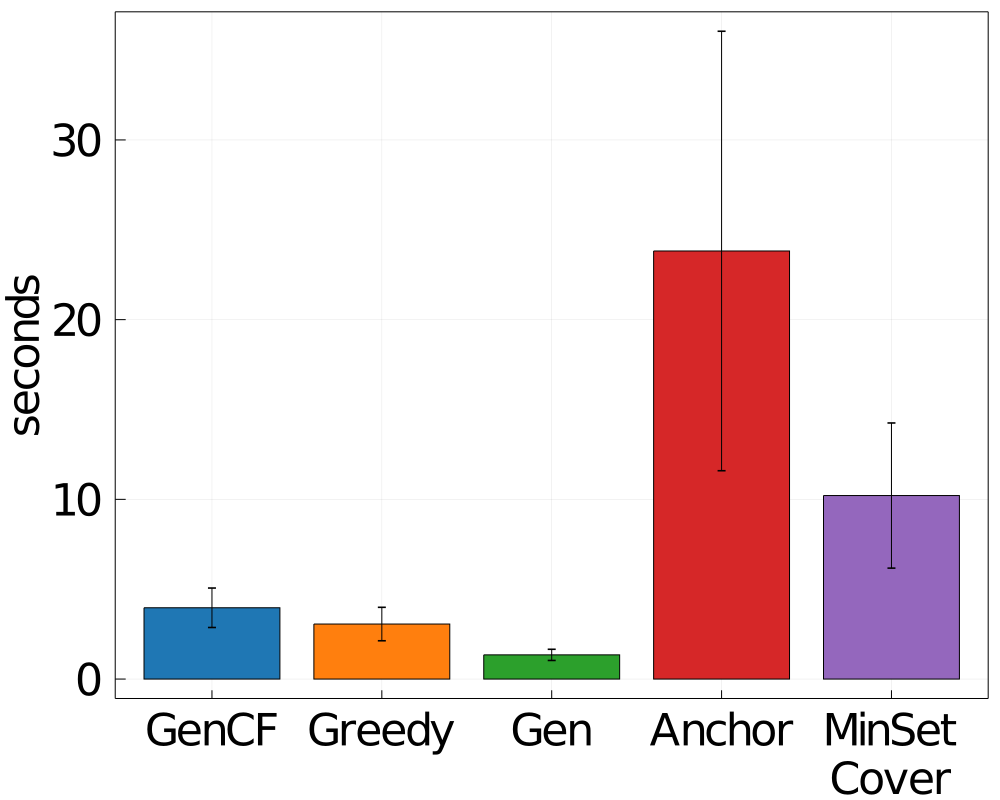}
\endminipage\hfill
\minipage{0.24\textwidth}
\centering\textbf{Cardinality of Classifier = 4}\par\medskip
  \includegraphics[width=\linewidth]{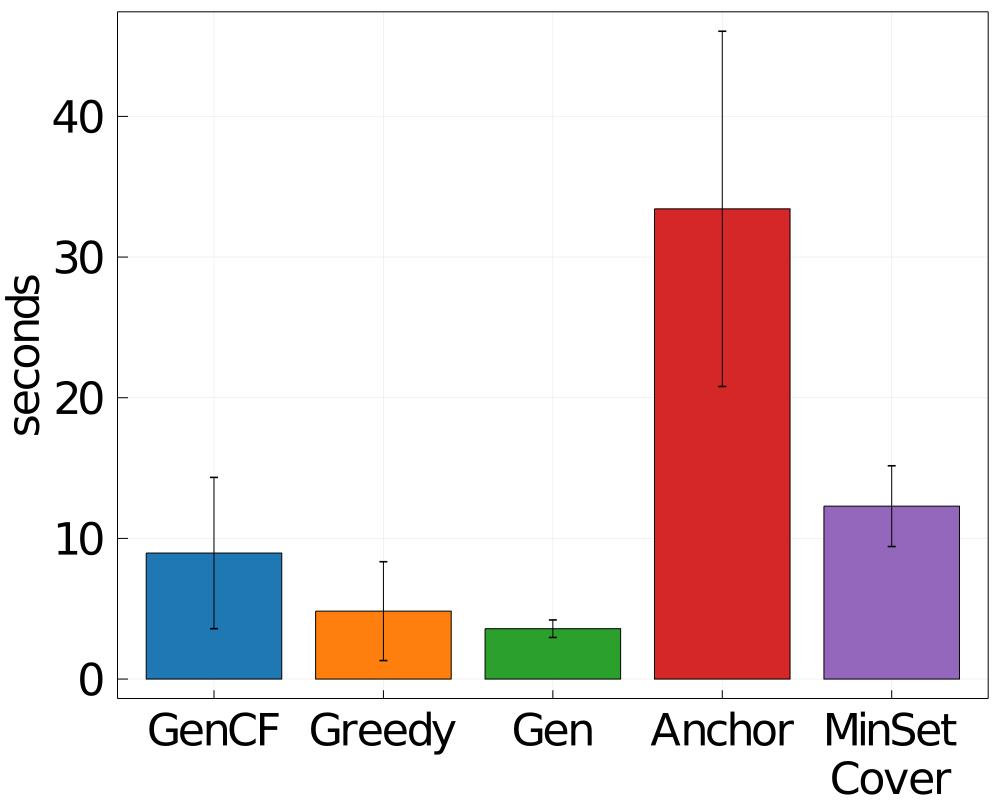}
\endminipage\hfill
\minipage{0.24\textwidth}
\centering\textbf{Cardinality of Classifier = 6}\par\medskip
  \includegraphics[width=\linewidth]{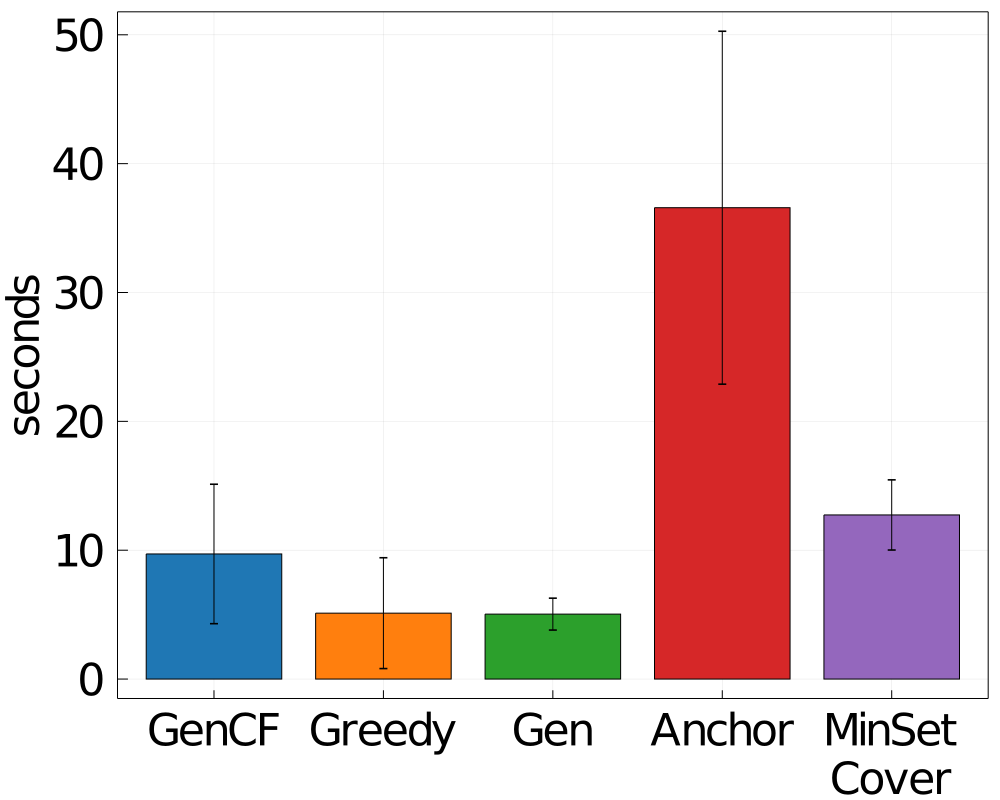}
\endminipage\hfill
\minipage{0.24\textwidth}%
\centering\textbf{Cardinality of Classifier = 8}\par\medskip
  \includegraphics[width=\linewidth]{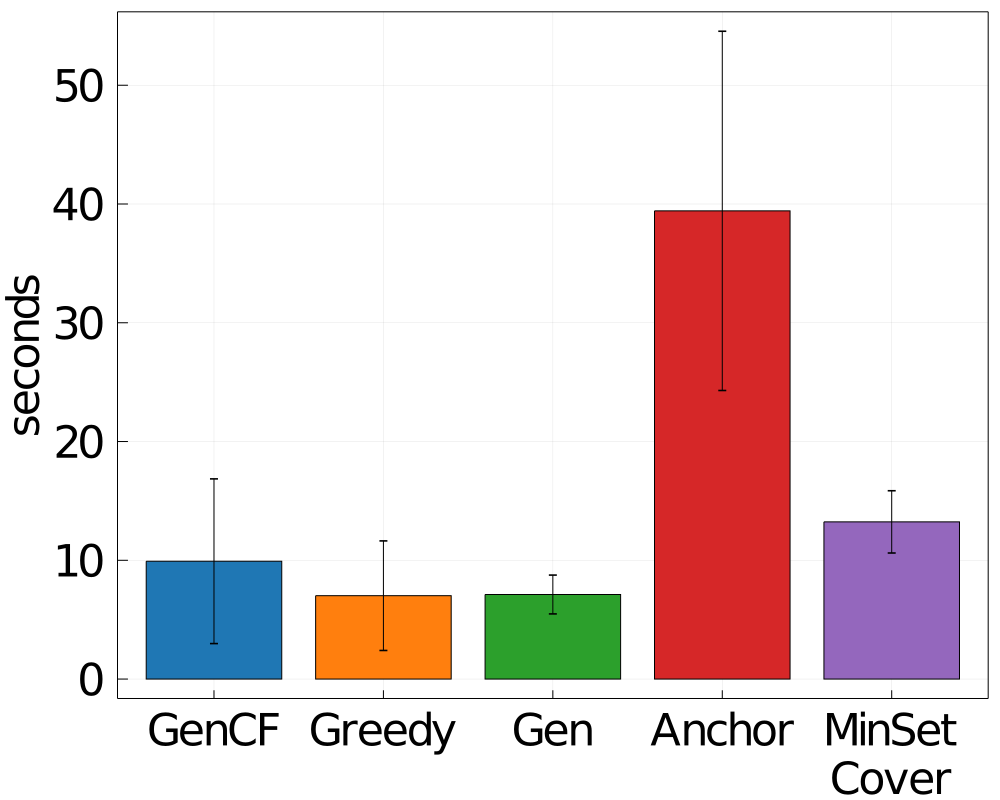}
\endminipage
\caption{ Comparison of runtime for \algogr\;(Gen), \algogrcf\;(GenCF), \algogreedy\;(Greedy), Anchor, and MinSetCover over 1000 synthetic classifiers with 2, 4, 6, and 8 rule components for the Credit dataset}.
\label{fig:ground_truth_runtime}
\end{figure*}

\begin{figure*}[htbp!]
\minipage{0.24\textwidth}
\centering\textbf{Credit Dataset}\par\medskip
  \includegraphics[width=\linewidth]{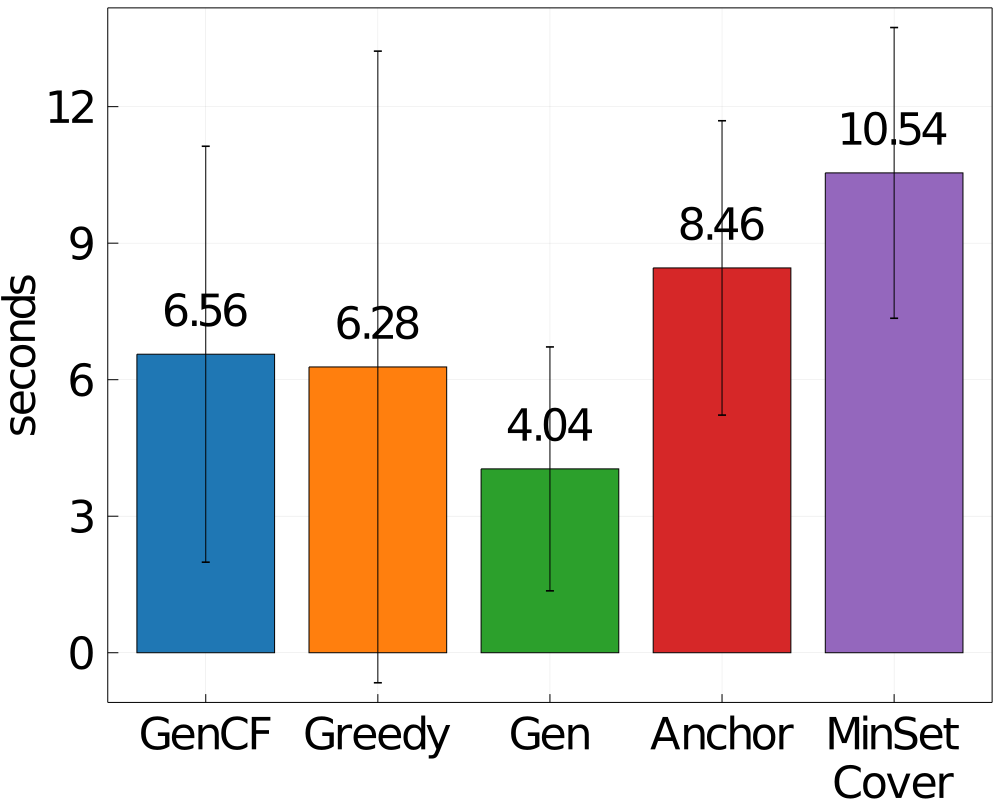}
\endminipage\hfill
\minipage{0.24\textwidth}
\centering\textbf{Adult Dataset}\par\medskip
  \includegraphics[width=\linewidth]{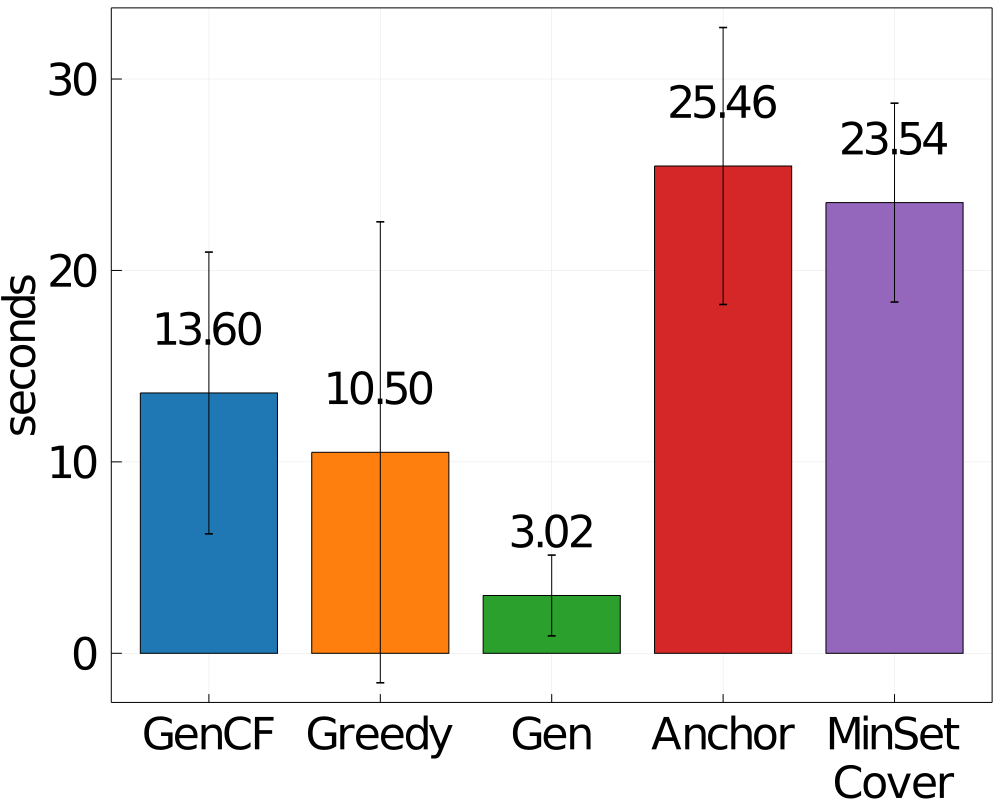}
\endminipage\hfill
\minipage{0.24\textwidth}%
\centering\textbf{Fico Dataset}\par\medskip
  \includegraphics[width=\linewidth]{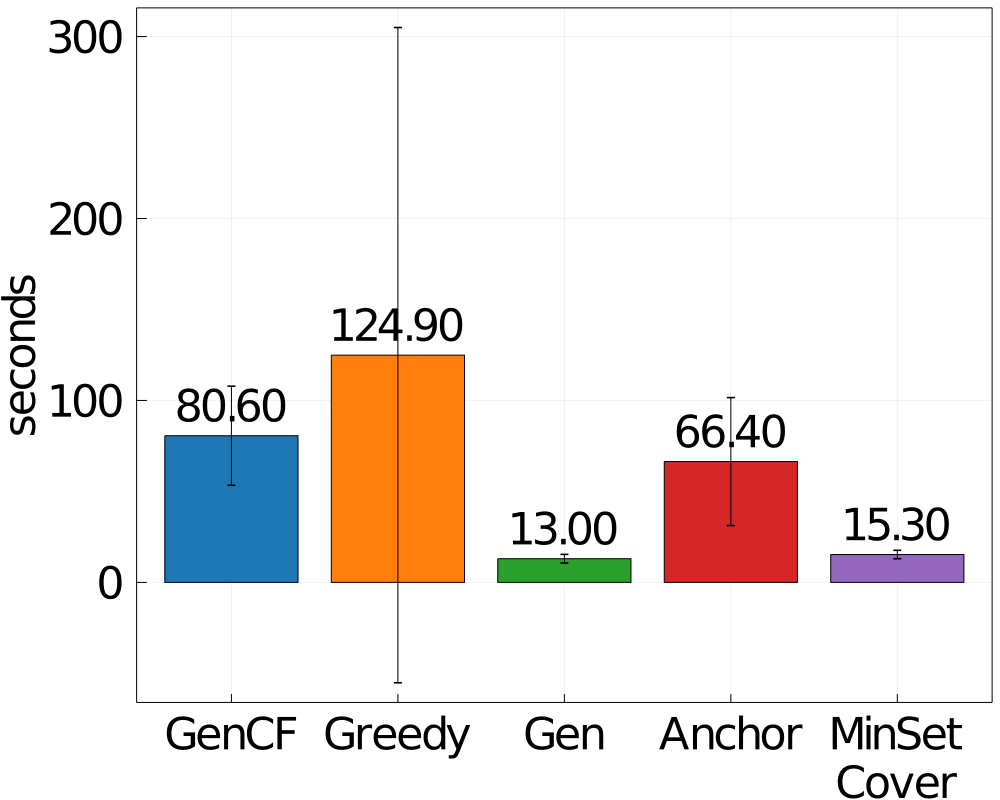}
 \endminipage\hfill
\minipage{0.24\textwidth}%
\centering\textbf{Yelp Dataset}\par\medskip
  \includegraphics[width=\linewidth]{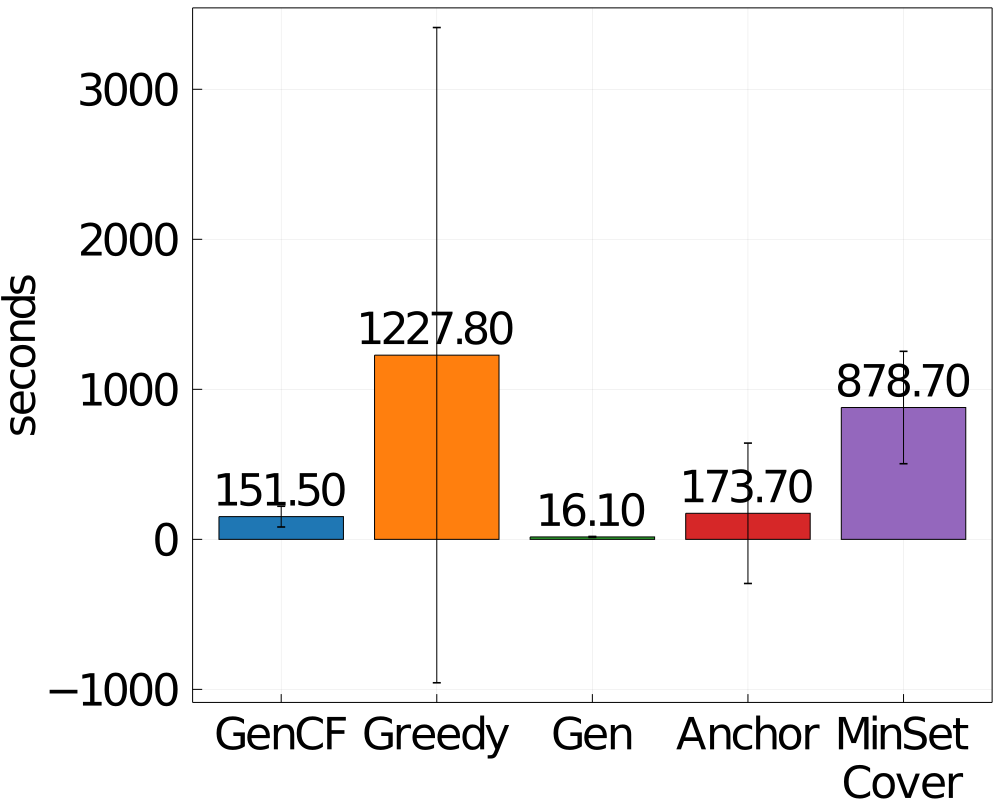}
\endminipage
\caption{Comparison of the average runtime (in seconds) of rules for \algogr, \algogrcf, \algogreedy, Anchor, and MinSetCover. We explain 10000 instances for the Credit, Adult, and Fico dataset, and 100 instances for the Yelp dataset.}
\label{fig:runtime_real_classifier}
\end{figure*}

\subsection{Runtime Comparison}
We measure the runtime of all considered algorithms for the synthetic and real classifiers. In particular, we investigate how the runtime is affected by the cardinality of the synthetic classifiers, as well as the sizes of the the different datasets.

\textbf{Synthetic classifiers.} Figure \ref{fig:ground_truth_runtime} shows the runtime of algorithms in synthetic classifiers with $2$, $4$, $6$, and $8$ rule components. 

\algogr, \algogrcf, and \algogreedy\, usually consume less time than Anchor and MinSetCover, regardless of the cardinality of rules behind the classifier. In particular, for the classifier with 8 rule components \algogr\, and \algogreedy\, are about  $5\times$ faster than Anchor and $1.8\times$ faster than MinSetCover; \algogrcf\, is about $3.9\times$ faster than Anchor and $1.3\times$ faster than MinSetCover.

We find that the larger the cardinality of the classifier rules, the longer the algorithms take to return a result. This is expected, since it takes more effort to build more complex rules. When the cardinality of classifier rules increases from $2$ to $6$, \algogr\, is more than 3$\times$ slower, \algogrcf\, is 2.8$\times$ slower, and \algogreedy\, is only 1.7$\times$ slower. \algogreedy\, is less affected by the increase in the cardinality. For instance, it takes almost the same time for classifiers with 4 and 6 rule components. This is because the algorithms can add several required rule components to a rule in every iteration with the counterfactual explanations, while in the traditional approach we can only add one more rule component to a rule in each iteration.

\textbf{Real Classifiers.} Figure \ref{fig:runtime_real_classifier} compares the runtime with real classifiers for each considered algorithm and dataset.

For the Credit dataset, the runtimes for all algorithms are similar, while \algogr\, is the fastest and MinSetCover is the slowest. Credit has $14$ variables  and thus the decision tree classifier is relatively small. This demonstrates that our \algogrcf\, and \algogreedy\, algorithms can efficiently generate consistent rule-based explanations without extra cost for a moderately complex datasets and classifiers.

For Adult, \algogr, \algogrcf, and \algogreedy\, are all significantly faster than Anchor and MinSetCover. For instance, \algogr\, is more than $6\times$ faster than Anchor and MinSetCover, whereas \algogreedy\, is more than $2\times$ faster. This performance difference can be explained by the fact that the dataset contains many variables that were one-hot encoded during preprocessing, which significantly increases the number features. Whereas Anchor and MinSetCover scale poorly in the number of features, our algorithms can treat one-hot encoded features as one feature. As a consequence, our algorithms can significantly outperform the existing systems in the presence of one-hot encoded variables. 

For Fico and Yelp, \algogrcf\, and in particular \algogreedy\, take much more time. This mainly because we use a strong verification mechanism in the two algorithms. The verification mechanism prevents the algorithms from stopping until they find a consistent rule, whereas the other algorithms might stop early and return inconsistent rules (c.f., Figure \ref{fig:breakdown_verified_real}). 

Another reason for the slower performance is the performance of the underlying classifier, which is particularly apparent for the Fico dataset. Since our algorithms use a counterfactual explanation system, \algogrcf\, and \algogreedy\, call the classifier significantly more often than Anchor and MinSetCover. Thus if the underlying classifier is slower, \algogrcf\, and \algogreedy\, are also much slower. Table~\ref{tab:classifier} presents the runtime for classifiers and GeCo on each dataset. For the Fico dataset, the classifier is more than 25$\times$ slower than that of the Credit dataset and 13 $\times$ slower than that of the Adult dataset. This led to GeCo taking 22.2$\times$ and 14.7$\times$ more time with the classifier of the Fico dataset compared to that of the Credit dataset and the Adult dataset. The increase of runtime in GeCo significantly increases the runtime of \algogrcf\, and \algogreedy as they rely on the GeCo to verify the rules. We will discuss more details in the microbenmarks in Sec~\ref{sec:Microbenchmarks}.

The Yelp dataset contains millions of instances and we use a significantly more complex classifier. For this reason, it is much harder for the algorithms to generate and verify rules. This is visible for our algorithms but also for MinSetCover, whose runtime is highly depend on the number of instances in the dataset. However, even for the slow classifier as on Fico Dataset, and large dataset as Yelp dataset, our \algogrcf can always generate rules in time that is at least comparable, and at times significantly faster, than Anchor and MinSetCover. This shows the power of using a genetic algorithm to build the rules with less run time.

In summary, \algogrcf\, can always finish in reasonable runtime to generate high quality rules regardless of the classifier speed or the size of the dataset. \algogreedy\, typically generates rules with the highest quality, and it is fast when the classifier and dataset have moderate size. For complex classifiers over large datasets, however, \algogrcf\, is more efficient than \algogreedy.

\begin{table}[t]
\centering
 \begin{tabular}{|c|c|c|c|c|} 
 \hline
    & Credit & Adult & Fico & Yelp \\ 
 \hline\hline
 Classifier Runtime & 0.0041 & 0.0079 & 0.1079 &  0.02081\\
 \hline
 GeCo Runtime & 0.0854 & 0.1294 & 1.9050 & 2.0793\\
 \hline
 \end{tabular}
 \vspace{1.2mm}
 \caption{Run time of the classifiers to predict 10,000 instances, and for GeCo to explain a single instance on each dataset.}
 \label{tab:classifier}
\end{table}

\begin{figure*}[htbp!]
\minipage{0.36\textwidth}
\centering\textbf{Runtime Breakdown}\par\medskip
  \includegraphics[width=\linewidth]{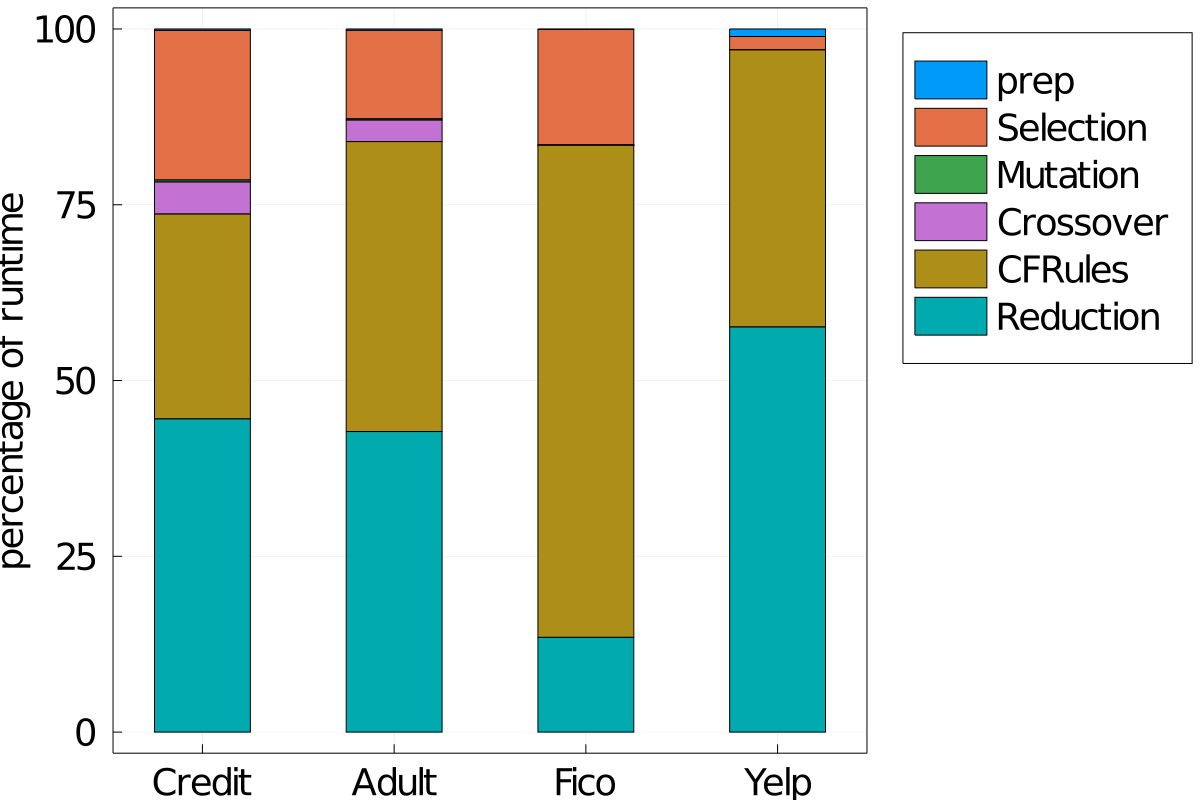}
  \caption{The break down of average run time into the main operators for \algogrcf\, algorithm.}
  \label{fig:break_down_time}
\endminipage\hfill
\minipage{0.30\textwidth}
\centering\textbf{Adult Dataset}\par\medskip
  \includegraphics[width=\linewidth]{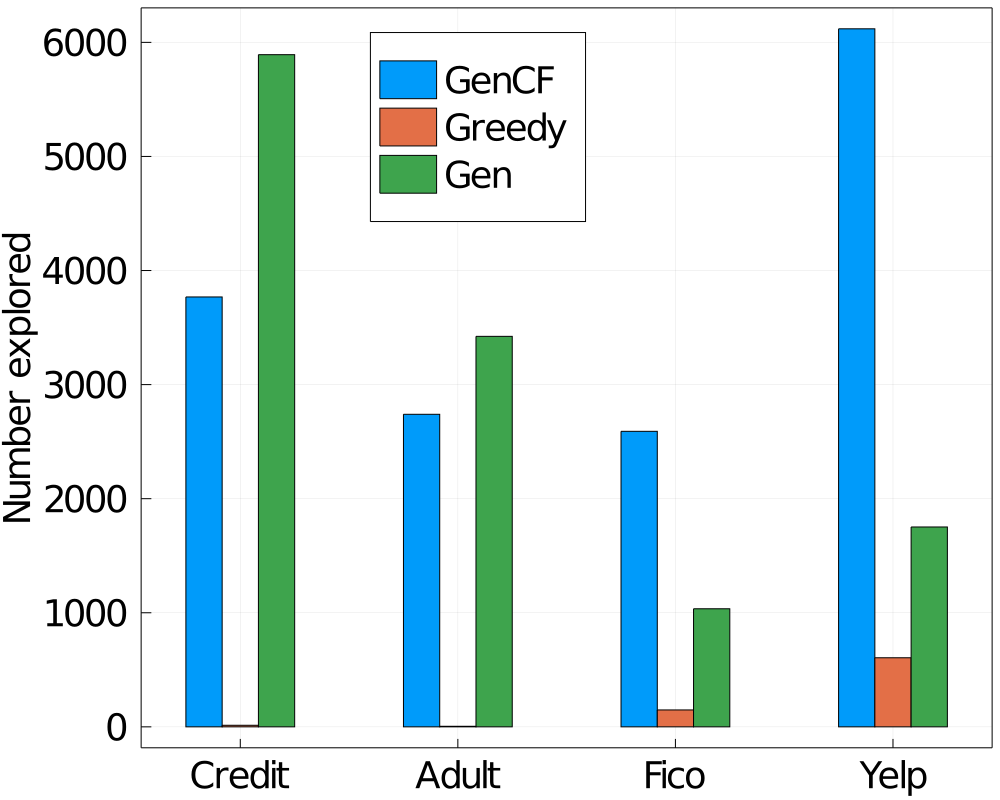}
  \caption{The number of rule candidates explored for \algogrcf\, , \algogr\, and \algogreedy.}
  \label{fig:break_down_explored}
\endminipage\hfill
\minipage{0.30\textwidth}%
\centering\textbf{Fico Dataset}\par\medskip
  \includegraphics[width=\linewidth]{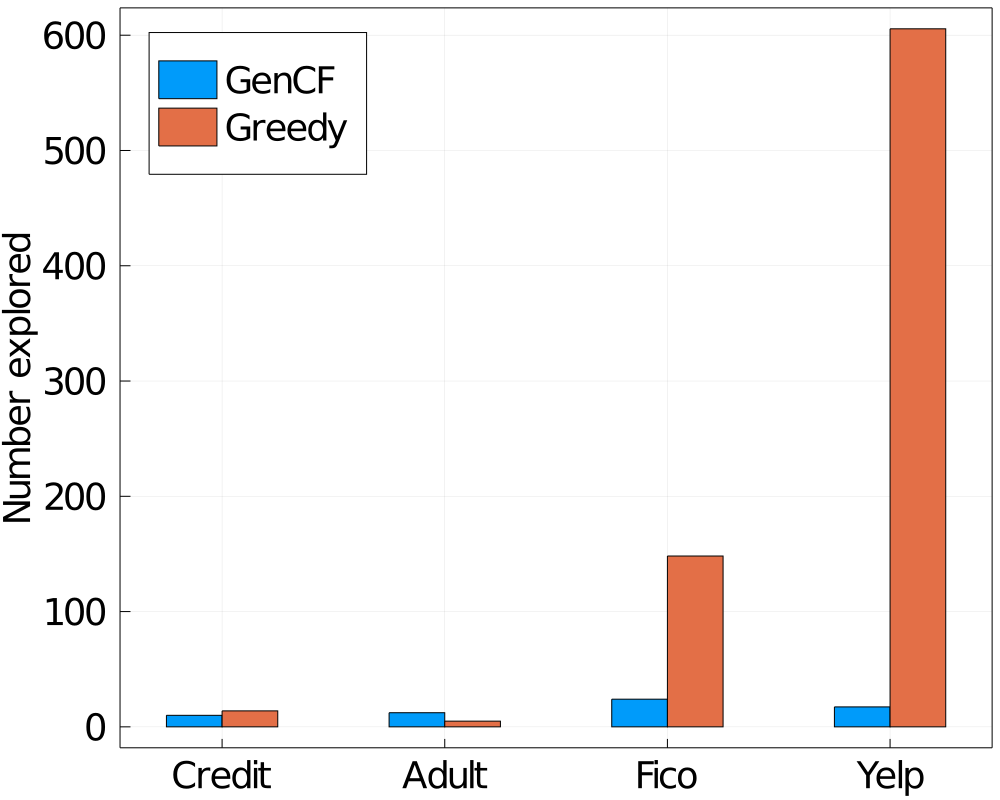}
  \caption{The number of rule candidates explored by GeCo for \algogrcf and \algogreedy.}
   \label{fig:break_down_explored_geco}
\endminipage
\end{figure*}

\subsection{Microbenchmarks}
\label{sec:Microbenchmarks}

In this section, we present the results for the microbenchmarks. We compare the runtime break down for each main operators for \algogrcf, the number of rule components explored for \algogrcf, \algogreedy\, and \algogr, and the number of rule components explored by the counterfactual explanation systems for \algogrcf\, and \algogreedy.  We evaluate our algorithms by explaining $100$ instances on Adult Dataset, Credit Dataset,  Fico Dataset and  $20$ instances on Yelp Dataset with corresponding classifiers.

\textbf{Breakdown of Runtime.}
Figure \ref{fig:break_down_time} presents the results for the runtime breakdown. We choose to not include \algogreedy\, since almost all of its runtime is from the counterfactual system. This is because \algogreedy\, only uses the counterfactual system, GeCo, to build and verify rules. Prep (i.e. Preparation) captures the runtime to compute the rule space and to build the initial population. Reduction captures the timed used to removed the redundant components from the returned rules using the counterfactual explanation systems. The runtime for selectFittest (Selection), Crossover, Mutate, and CFRules is accumulated over all iterations.

As discussed in Section~\ref{sec:genetic_rule_with_gego}, we only run the counterfactual explanation once in the function \textbf{CFRules} for each rule candidate to optimize the performance. This also gives us the information we need for function \textbf{consistentCF} which used in selectFittest. Therefore, we do not run counterfactual explanations in the selectFittest and the time used by function \textbf{CFRules} captures the time used by the counterfactual system in building the rules. And the total time used by the counterfactual system is the sum of the runtime in CFRules and Reduction.

The results show that the CFRules and Reduction are the most time-consuming operations. This is not surprising since these two operations rely on the counterfactual system, which is costly. And thus, the majority of the runtime is consumed by the counterfactual system. If we can reduce the runtime for the underlying counterfactual system, we would like to see a huge gain in the runtime of our algorithms.

\textbf{Number of Candidate Rules Explored.}
Figure \ref{fig:break_down_explored} shows the number of candidates explored for each of the algorithms. The result shows that our optimizations in \algogrcf\, and \algogreedy\, effectively limit the total number of rules explored in the Credit and Adult datasets when the classifier is moderately complex and guide \algogrcf\, and \algogreedy\, to search candidates that are more likely to be consistent. When the classifier is complex as in the Fico and Yelp datasets, our optimizations prevent \algogrcf\, and \algogreedy\,  from stopping early with inconsistent rules and push the algorithms to explore more rules until finding a real consistent one. 

\textbf{The Number of GeCo Runs.}
Figure \ref{fig:break_down_explored_geco} presents how many times we use GeCo in each of the algorithms in the four datasets. For moderate dataset and classifiers, like Adult and Credit, the number if rules explored by \algogrcf\, and \algogreedy\, are similar. However, when the datasets and classifiers become large and complex, like Fico and Yelp, the genetic algorithm in \algogrcf\,  significantly reduces the number of runs using the counterfactual systems (6 times fewer in the Fico dataset and 35 times fewer in the Yelp dataset). This explains why the \algogrcf\, is more efficient than \algogreedy\, for large datasets with complex classifiers.

\balance


\section{Limitations and Future Work}
We have illustrated the effectiveness and efficiency of using the underlying counterfactual explanation model to generate rule-based explanations and compared it with other state-of-the-art algorithms. Now we come to its limitations and opportunities for future work.

\textbf{Bound of Rule Components.} In our algorithms, to reduce the search space of the rules, we limit our rules to strictly related to the values of the input instance, and the bound of any rule components must be the corresponding feature value. That is, our rule components can only be larger, smaller, or equal to the feature value. 
However, the range of the rules can be broader. 
For example, there is an input instance $x = \{F_1 = 3\}$ and the classifier $C$ has a rule $F_1 < 10$. 
Since we use the feature value $F_1$ as the bound of the rule component, we output the rule as $F_1 < 3$, which is narrower than the real rule. 
Currently, we want to analyze the behavior of the classifier with respect to the input instance, so this strict bound satisfies our expectations. 
In the future, we may want to take advantage of this strict bound to make the rule more general. 

\textbf{Realistic Feature Value Distributions}. In GeCo and the sampling process of our algorithms (and many other state-of-the-art Counterfactual Explanation models), we assume the perturbation distributions as the instance search space. 
This is sufficient to leverage the behavior of the model which generates interpretable explanations. 
However, how to estimate such distributions is still a questionable and challenging problem, such as how to represent the causal dependency between different features. 
Designing ways to find such distributions will benefit multiple explanation methods.

\textbf{Underlying Counterfactual Explanation System}. In the experiments, we find that stability and run time of our algorithm is highly depend on the underlying counterfactual system. In our implementation, we use GeCo as our underlying Counterfactual Explanation System, which is currently the best counterfactual explanation system we found. However, we observed that GeCo can still be costy and unstable in extreme cases, which negatively affects the run time and stability of our systems. If there is a more efficient and stable counterfactual explanation system, we would expect a huge gain in our systems.

\textbf{Better Counterfactual Explanation Model}. In this paper, we use the counterfactual explanation model to generate rule-based explanations. 
Similarly, we can also use the rule-based explanation to help identify which features needed to be changed for counterfactual explanations. If there is a well-established rule-based explanation model, we can apply the idea to facilitate the counterfactual explanation model.

\textbf{Static Data and Classifier.}
Currently, our rule-based explanation algorithms assume the underlying data and classifier to be static. 
Therefore, our algorithms are subject to changes in the data and classifier. 
We plan to explore how we can generate explanations that are robust to small changes in the data distribution or classifier. 
This is related to the more general problem of robust machine learning.

\balance

\section{Conclusion}

Rule-based explanations are highly desirable for automated, high
stakes decisions, yet they are computationally intractable.  In this
paper we have described a new approach for computing rule-based
explanations, which uses counterfactual explanation system as an
oracle.  We also use the counterfactual explanation system to robustly
verify the global consistency of the rules.  We have described a base
genetic algorithm (\algogr) and two extended algorithms (\algogrcf\,
and \algogreedy\,) that integrate the results from the counterfactual
explanations in order to build globally consistent, informative
rule-based explanations.  We conducted an extensive experimental
evaluation, proving that the rule-based explanations returned by our
system are globally consistent, and have fewer rule components,
i.e. are more informative, than those returned by other systems
described in the literature.

\balance

\begin{acks}
  This work was partially supported by NSF IIS 1907997 and NSF-BSF
  2109922.
\end{acks}

\clearpage
\balance
\bibliographystyle{ACM-Reference-Format}
\bibliography{references}

\appendix
\section{Appendix: Choosing the underlying Counterfactual Explanation model}
\label{sec:choose_cf}
Both Genetic Rule with CF algorithm and CF Greedy algorithm build on top of a Counterfactual Explanation model and the performance of these two algorithms relies heavily on the reliability and efficiency of the Counterfactual Explanation model. Therefore, it is crucial for us to find a Counterfactual Explanation model that can support our needs. In this section, we talk about how we choose GeCo as our underlying Counterfactual Explanation model.

\subsection{Requirements}
Here, we illustrate what our algorithms need from the underlying Counterfactual Explanation model, which gives us direction to decide which Counterfactual Explanation model to select.

\textbf{Flexible Feasible Constraints.} No matter whether building up the rules using CF or validating rules using CF, we have to first convert the rules to the Feasible Constraints before running the Counterfactual Explanation model. This is how we link our Rule-based model to the Counterfactual Explanation model.  Also, we don't want to fix the Feasible Constraints when we import the dataset/classifier, but we want to change the Feasible Constraints corresponding to the rules each time we run the Counterfactual Explanation model. Therefore, Counterfactual Explanation model should support Feasible Constraints, and also these Feasible Constraints should be flexible and not static per dataset/classifier. Since these Feasible Constraints are generated by code and change all the time, ideally we also want it to have low overhead and a friendly interface to generate the Constraints.

\textbf{Black-box Algorithm.} Since our Rule-based algorithms support different kinds of classifiers and treat them as black-box, we should not break our guarantee by the underlying Counterfactual Explanation model. Therefore, it should also support different types of classifiers and should be black-box algorithm with respect to the classifier.

\textbf{Reliability.} 
Our Rule-based algorithms rely heavily on the result from the Counterfactual Explanation model and we blindly think that the results are correct. Particularly, we assume that (1) the Counterfactual Explanations returned with the correct label should be valid Counterfactual Explanations and (2) it should always return at least one valid Counterfactual Explanation when there are some valid Counterfactual Explanations under the constraints. We use the first one to build our rules as function \textbf{CFForRule}. And we use the second one to verify whether the rule is sound or not in function \textbf{CFForRule}. Both of the functions are used in Algorithm \ref{alg:genetic_rule_geco} and \ref{alg:geco_greedy}. Therefore, the Counterfactual Explanation model must match the assumptions we make.

\textbf{Effectiveness.} 
As discussed, we implement the function \textbf{CFForRule} based on the idea in Section \ref{sec:convert_counterfactual_to_rules}. And in order to make sure our rule is interpretable (with low cardinality), we also want to make sure that the Counterfactual Explanation generated has no useless feature change. Otherwise, we may introduce useless rule components, which is what we want to avoid. By useless feature change, we mean that if substitute a feature value in Counterfactual Explanation that is different from the original instance to the corresponding feature value in the original instance, the new instance should always be labeled as "undesired" (if we removed the feature change, it will be undesired again). Therefore, we want the Counterfactual Explanation model to generate Counterfactual Explanations without useless feature changes. In real measurements, this is similar to how close the Counterfactual Explanations are to the original instance.

\textbf{Efficiency.} 
We use the Counterfactual Explanation model quite often in our algorithms. We use it each time we may want to generate new rules or verify rules. Therefore, although run-time may not be a required factor, we want to use a Counterfactual Explanation model that can finish in a reasonable time to make sure that our run-time is reasonable.

\subsection{Counterfactual Explanation model}
We considered 13 different Counterfactual Explanation models in literature and we briefly discuss them as below:

\textbf{AR}. AR\cite{AR} or Actionable Recourse is suggested by Ustun et al. to generate minimal cost actions to flip a "undesired" instance to "desired" instance. It simulated an approximate linear model of the
classifications and use integer programming to find results with minimum costs (an optimization problem). It supports user–specified constraints as actionable or immutable features.

\textbf{CCHVAE}. CCHVAE\cite{CCHVAE} is provided by Pawelczyk et al., which use VAE to approximate the classifier structure and find counterfactual in the high possible region.  It doesn't support the Feasible Constraints.

\textbf{CEM}. CEM\cite{CEM} is proposed by Dhurandhar et al., which finds minimum-cost counterfactual instances with an elastic–net regularization. It doesn't support the Feasible Constraints.

\textbf{CLUE}. CLUE\cite{CLUE} is from Antorán et al., which considers the classifier’s uncertainty and uses VAE to approximate a generative model to search CFs towards low uncertainty space. It supports the Feasible Constraints respect to the data set. 

\textbf{CRUDS}. CRUDS\cite{cruds} is suggested by Downs et al., which uses Conditional and Disentangled Subspace Variational Autoencoder to search counterfactuals over the interested latent features. It supports user–specified constraints.

\textbf{DICE}. Dice\cite{dice} is provided by Mothilal et al., which uses gradient descent to solve the optimization problem that is a trades-off between the diversity and the distance from the original distance. It supports the Feasible Constraints with feature ranges and immutability features.

\textbf{FACE}. FACE\cite{face} is from Poyiadzi et al. who uses graphs to find the shortest path in the high–density regions. It only considers the result in the dataset and supports the Feasible Constraints. There are two variants that differ by the graphs (FACE–EPS with epsilon–graph and FACE–KNN with knn-graph).

\textbf{Growing Spheres (GS)}. Growing Spheres \cite{gs} is proposed by Laugel et al., which utilizes a random search algorithm to generate a sample around the original instance with growing hyperspheres until finding a counterfactual. Feasible constrain is supported.

\textbf{FeatureTweak}. FeatureTweak\cite{ft}  is provided by Tolomei et al., which tweaks input features over a designed Random Forest classifier. It mainly works for the tree-based classifiers and doesn't support Feasible Constraints.

\textbf{FOCUS}. FOCUS\cite{focus} is from Lucic et al., which uses probabilistic model approximations of the original tree ensemble and gradient-based optimization to find the counterfactual close to the original instance. It doesn't support user-specified constraints and only work for tree-based classifiers.

\textbf{REVISE}. REVISE\cite{revise} is suggested by Joshi et al., which uses a
variational autoencoder (VAE) to estimate the generative model and latent space to find the counterfactual. It doesn't support Feasible constrain.

\textbf{Wachter}. Wachter\cite{Wachter} is from Wachter et al., which uses gradient descent to solve the optimization problem of minimizing the 1-norm distance from the counterfactual to the original instance. It doesn't support Feasible constrain.

\textbf{GeCo}. GeCo \cite{geco} is proposed by Schleich et al., which utilizes a heuristic random search algorithm to generate counterfactual closed to the original instance and with few feature changes. It supports feasible constrain for immutable features, specific feature ranges and correlated features.

\begin{table*}[htbp!]
\centering
 \begin{tabular}{|c|c|c|c|c|c|c|c|c|} 
 \hline & 
 \multicolumn{4}{c|}{Artificial Neural Network} & \multicolumn{4}{c|}{Logistic Regression}\\
 \hline
    & Redund. & Viol. & Success & Time (s) &  Redund. & Viol. & Success & Time (s) \\ 
 \hline\hline DICE& 0.50 & 0.10&\textbf{1.00} &0.13&0.26&0.10&\textbf{1.00}&0.12\\
 \hline AR& \textbf{0.00}&\textbf{0.00} & 0.29&2.07&\textbf{0.00}&\textbf{0.00}&\textbf{1.00}&49.06\\
\hline CCHVAE& NaN&NaN&0.00&2.65&4.16&1.35&\textbf{1.00}&1.14\\
\hline CEM&3.96&\textbf{0.00}&\textbf{1.00}&0.60&4.34&0.06&\textbf{1.00}&0.53\\
\hline CLUE&7.79&1.26&\textbf{1.00}&1.55&NaN&NaN&0.00&1.42\\
\hline CRUDS&11.81&1.31&0.42&4.80&11.83	&1.35&\textbf{1.00}&4.16\\
\hline FACE\_KNN&4.97&1.42&\textbf{1.00}&6.10&3.94&1.37&0.99&6.15\\
\hline FACE\_EPS&5.15&1.47&0.98&6.27&3.89&1.43&0.74&6.31\\
\hline GS&3.82&\textbf{0.00}&\textbf{1.00}&\textbf{0.01}&3.37&\textbf{0.00}&\textbf{1.00}&\textbf{0.01}\\
\hline REVISE&NaN&NaN&0.00&7.40&10.56&1.22&0.92&7.32\\
\hline WACHTER&4.44&1.0&0.50&15.52&1.10&1.00&\textbf{1.00}&0.02\\
\hline GeCo&\textbf{0.00}&\textbf{0.00}&\textbf{1.00}&0.66&\textbf{0.00}&\textbf{0.00}&\textbf{1.00}&0.53\\
 \hline
\end{tabular}
\vspace{1.2mm}
 \caption{Results from Carla benchmark on 12 Counterfactual Explanation Systems for Adult dataset with Artificial Neural Network and Logistic Regression classifiers using Tensorflow and Pytorch frameworks.}
 \label{tab:ANN_LR_table}
\end{table*}

\begin{table*}[htbp!]
\centering
 \begin{tabular}{|c|c|c|c|c|c|c|c|c|} 
 \hline & 
 \multicolumn{4}{c|}{Artificial Neural Network + SKlearn} & \multicolumn{4}{c|}{Logistic Regression + XGboost}\\
 \hline
    & Redund. & viol. & Success & Time (s) &  Redund. & Viol. & Success & Time (s) \\ 
 \hline\hline FOCUS&4.04&1.00&1.00&0.15&4.04&1.00&1.00&0.33\\
 \hline FEATURE\_TWEAKING&0.99&0.37&0.45&\textbf{0.00}&\textbf{0.00}&\textbf{0.00}&0.81&\textbf{0.02}\\
 \hline GeCo&\textbf{0.00}&\textbf{0.00}&\textbf{1.00}&0.10&\textbf{0.00}&\textbf{0.00}&\textbf{1.00}&0.14\\
 \hline
\end{tabular}
\vspace{1.2mm}
 \caption{Results from Carla benchmark on 3 Counterfactual Explanation Systems for Adult dataset with Artificial Neural Network on SKlearn and Logistic Regression classifiers on XGboost.}
 \label{tab:sX_table}
\end{table*}

\begin{figure*}[htbp!]
\includegraphics[width=\linewidth]{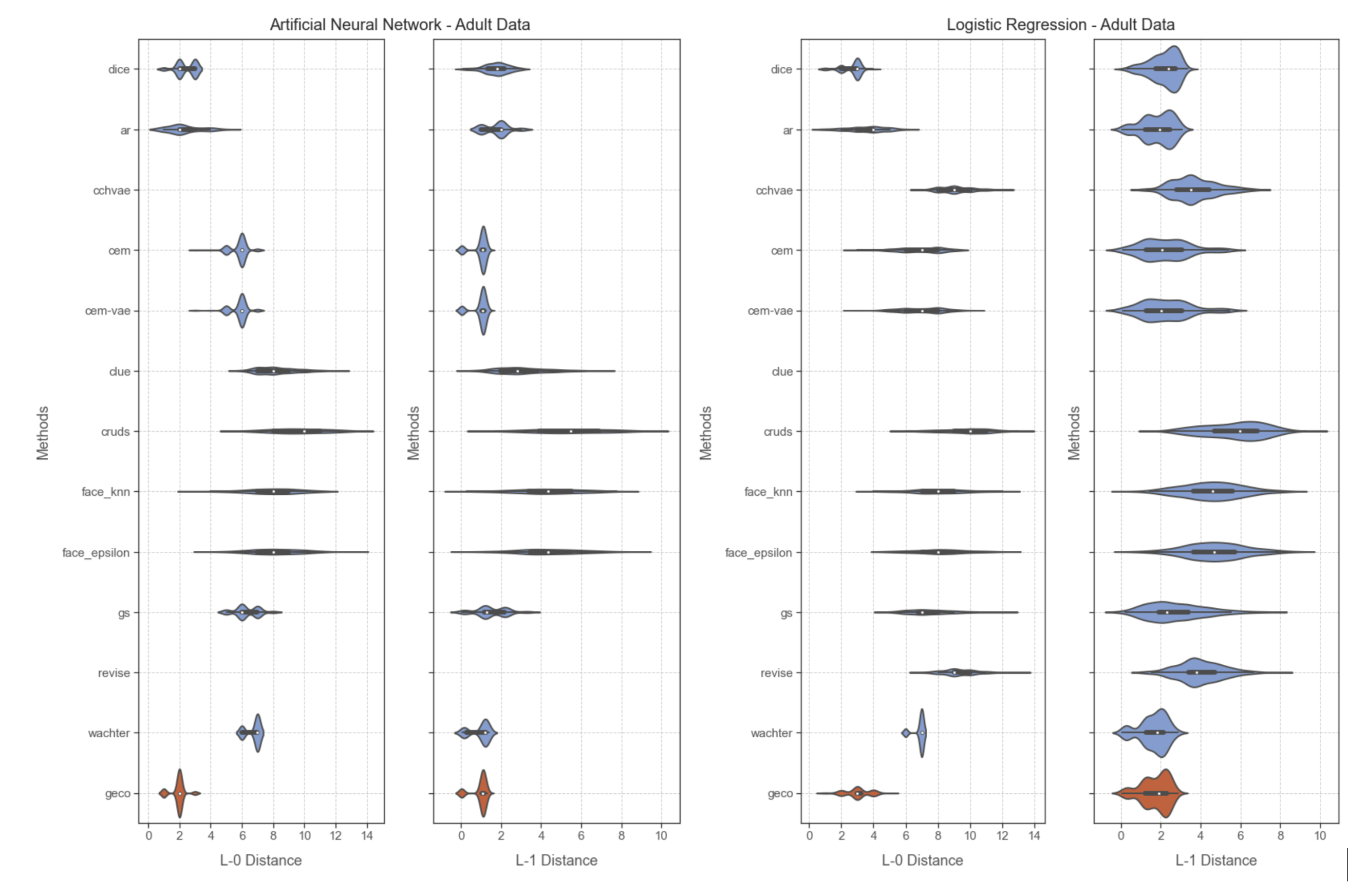}
\caption{L0 and L1 distances from Carla benchmark on 12 Counterfactual Explanation Systems for Adult dataset with Artificial Neural Network and Logistic Regression classifiers using Tensorflow and Pytorch frameworks.}
\label{fig:ANN_LR_G}
\end{figure*}

\begin{figure*}[htbp!]
\includegraphics[width=\linewidth]{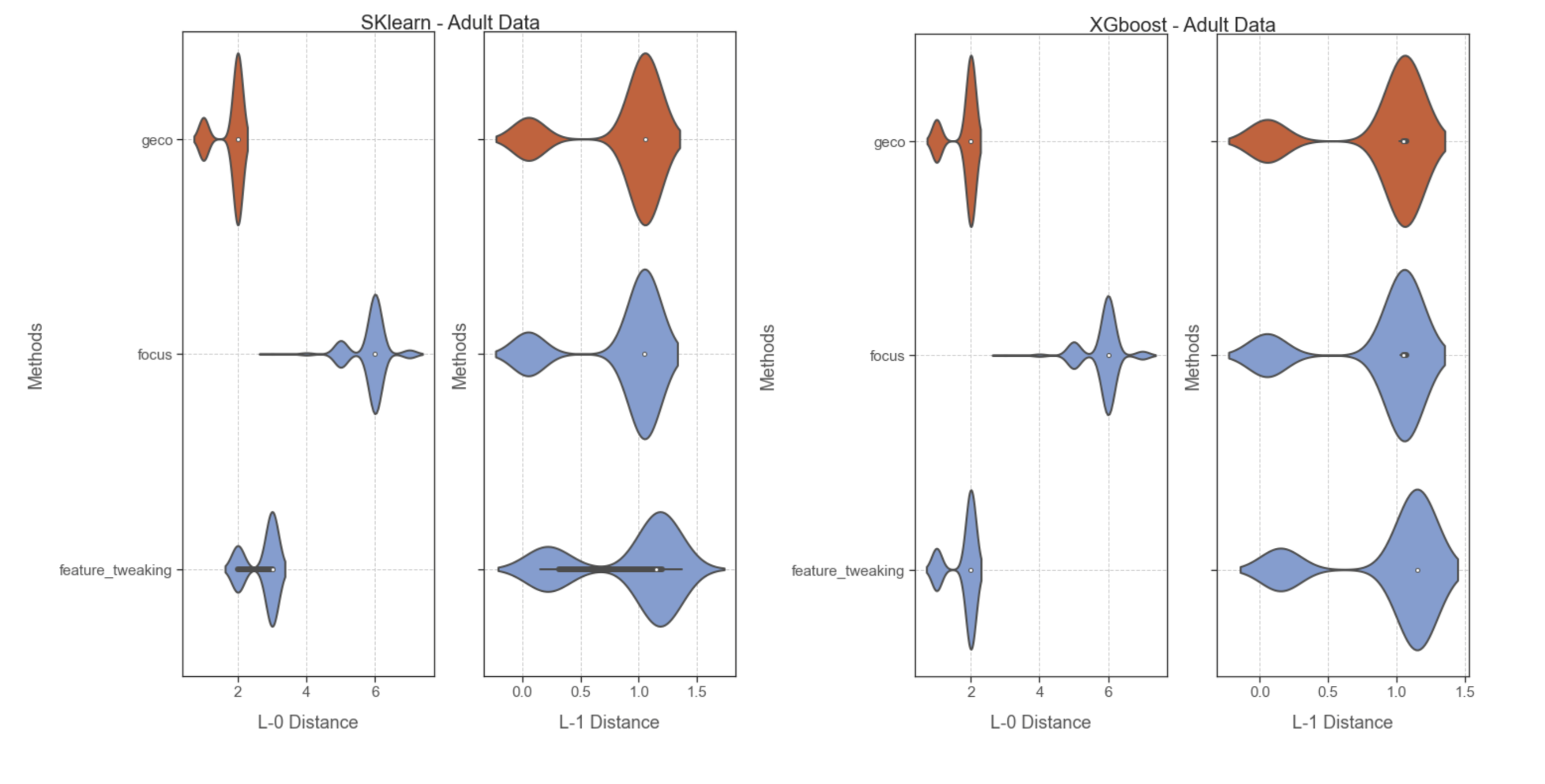}
\caption{L0 and L1 distances from Carla benchmark on 3 Counterfactual Explanation Systems for Adult dataset with Artificial Neural Network on SKlearn and Logistic Regression classifiers on XGboost.}
\label{fig:sX_graph}
\end{figure*}

\subsection{Carla as the Comparison Benchmark}
Carla \cite{carla} is a library that is used to benchmark Counterfactual Explanations on common datasets and classifiers. We use Carla as the benchmark to compare 13 different Counterfactual Explanation models (in total 15 variations) discussed above. We run the benchmark on the Adult dataset \cite{adult} on the provided ANN (Artificial Neural Network with 2 hidden layers and ReLU activation function) classifier and LR (Linear Model with no hidden layer and no activation function) classifier on Tensorflow, Pytorch, SKlearn, and XGBoostML frameworks. For these four frameworks, only GeCo supports all of them. Focus and FeatureTweak only support SKlearn and XGBoostML, while all others except GeCo don't support these two. Therefore, we run GeCo, Focus and FeatureTweak on the SKlearn and XGBoostML frameworks, and run GeCo with all other systems on Tensorflow and Pytorch frameworks. 

The results from Carla are shown in Table \ref{tab:ANN_LR_table} and Figure \ref{fig:ANN_LR_G} for Tensorflow and Pytorch frameworks and Table \ref{tab:sX_table} and Figure \ref{fig:sX_graph} for SKlearn and XGBoostML frameworks. 

Carla returns six parameters that we are interested in: (1) Average redundant feature number (redund) represents the number of useless feature changes and our Rule-based models want this to be as small as possible. (2) Average violations (viol) represent the number of features that we set to be immutable but the generated counterfactual changes. Average violations are not equal to zero means that the system doesn't support well for Feasible Constraints. Our Rule-based models want this to always be zero. (3) Success represents the rate we can successfully generate Counterfactual Explanations. Our Rule-based models want this to always be one (always success). (4) Time represents the average run-time of the Counterfactual Explanation System to generate Counterfactual for one instance. Ideally, our Rule-based models want this to be small. (5) l0 distance represents the average number of feature changes per counterfactual and this number should be small as possible. (6) l1 distance represents how close the counterfactual is to the original distance.

As discussed above, the ideal Counterfactual Explanation System should support the Feasible Constraints, support all types of classifiers and machine learning frameworks, should always return valid counterfactuals if possible, should not return redundant/useless feature changes, and should finish in a reasonable time. By looking at the result, only AR, GS, and \textbf{GeCo} support Feasible Constraints that do not violate the immutable features (rules). Only AR and \textbf{GeCo} return Counterfactual Explanations without redundancy. Only Dice, CEM, GS, Focus, and \textbf{GeCo} always return Counterfactual Explanations. Surprisingly, only GeCo fulfills all the requirements we want. Even though its run-time is not the best, it can finish in less than a second, which should be reasonable for us to use. Therefore, we choose GeCo as our underlying Counterfactual Explanation System.
\end{document}